\documentclass[journal]{IEEEtran}
\usepackage{underscore}
\usepackage{cite}
\usepackage{amsmath,amssymb,amsfonts}
\usepackage{algorithmic}
\usepackage{graphicx}
\usepackage{textcomp}
\usepackage[table]{xcolor}
\usepackage{booktabs}
\usepackage{multirow}
\usepackage{array}
\usepackage{tikz}
\usetikzlibrary{shapes,arrows.meta,positioning,fit,backgrounds,calc}
\usepackage{url}
\usepackage[hidelinks]{hyperref}
\usepackage{tabularx}
\usepackage{makecell}

\definecolor{myblue}{HTML}{1B1B5C}

\newcommand{\graphG}{\mathcal{G}}

\newcommand{\nodeset}{\mathcal{V}}
\newcommand{\edgeset}{\mathcal{E}}
\newcommand{\eventset}{\mathcal{O}}

\usepackage{amsthm}
\theoremstyle{definition}
\newtheorem{example}{Example}[section]
\newtheorem{agexample}[example]{Agent-Evolution Example}

\usepackage[most]{tcolorbox}

\definecolor{TakeawayBrown}{HTML}{00007B} 
\definecolor{OpenGapRed}{HTML}{9E4F4F}
\definecolor{BoxBack}{HTML}{FFFDFC}

\usepackage{fontawesome5}
\newtcolorbox{insightbox}{
    enhanced,
    breakable,
    colback=blue!3,
    colframe=blue!3,
    boxrule=0pt,
    frame hidden,
    sharp corners,
    borderline west={3pt}{0pt}{blue!85!black},
    left=13pt,
    right=11pt,
    top=5pt,
    bottom=5pt,
    boxsep=0pt,
    before skip=6pt,
    after skip=6pt
}

\definecolor{AcademicSlate}{RGB}{55, 80, 105}

\newtcolorbox{takeawaybox}{
    enhanced,
        breakable,
    colback=AcademicSlate!3,
    colframe=AcademicSlate!3,
    boxrule=0pt,
    frame hidden,
    sharp corners,
    borderline west={3pt}{0pt}{AcademicSlate!80!black},
    left=13pt,
    right=11pt,
    top=5pt,
    bottom=5pt,
    boxsep=0pt,
    before skip=6pt,
    after skip=6pt
}

\begin{document}

\title{Self-Evolving Agents as Dynamic Graph Transformation: A Survey and New Perspective}

\author{Yuanyuan~Xu,
Wenjie~Zhang, Yin~Chen, Xuemin~Lin~\IEEEmembership{Fellow,~IEEE}, and Ying~Zhang
\thanks{Manuscript received \today.
Corresponding author: Ying Zhang (email: ying.zhang@zjgsu.edu.cn).}
\thanks{Yuanyuan Xu and Wenjie Zhang are with the School of Computer Science and Engineering, The University of New South Wales, Sydney, NSW 2052, Australia~(e-mail: yuanyuan.xu@unsw.edu.au; wenjie.zhang@unsw.edu.au).

Yin Chen is with the Faculty of Engineering and Information Technology, University of Technology Sydney, NSW 2007, Australia~(e-mail: yin.chen@student.uts.edu.au).

Xuemin Lin is with the School of Data Science, The Chinese University of Hong Kong-Shenzhen, Shenzhen 518172, China~(e-mail: xuemin.lin@gmail.com).

Ying Zhang is with the Laboratory for Statistical Monitoring and Intelligent Governance of Common Prosperity, Zhejiang Gongshang University, Hangzhou, Zhejiang 310018, China~(e-mail: ying.zhang@zjgsu.edu.cn).
}}

\markboth{IEEE Transactions on Knowledge and Data Engineering, Vol.~XX, No.~X, 2026}%
{Xu \MakeLowercase{\textit{et al.}}: Agent Evolution as Dynamic Graph Transformation: A Survey}

\maketitle

\begin{abstract}
Large language model (LLM)-based agents are increasingly becoming self-evolving systems that persist across interactions, maintain memories, use tools, acquire skills, refine workflows, and coordinate with other agents. These capabilities make agent states structural and dynamic: entities, relations, attributes, dependencies, and execution structures change with new evidence, feedback, and environmental conditions. Existing graph-agent surveys typically treat graphs as support structures for agent functions rather than as evolving substrates, while self-evolving-agent surveys focus on agent-level mechanisms and rarely discuss graph topology evolution. Thus, the coupling between evolving agent state and dynamic graph topology remains underexplored. This survey connects these two research lines by framing \textit{agent evolution as dynamic graph transformation}. We model agent state as a dynamic graph, where memories, tools, skills, workflows, and inter-agent relations are represented as typed nodes, edges, and subgraphs updated through schema-constrained rewrites. Based on this formulation, we organize existing dynamic-graph-based methods for self-evolving agents into four taxonomies: node/feature evolution, edge/topology evolution, subgraph activation, and cross-component co-evolution. Building on this taxonomy, we propose dynamic graph learning as reusable infrastructure for self-evolving agents and map nine dynamic-graph-learning subfields to agent-evolution capabilities, discussing their adaptations and possible failure modes. Finally, we discuss five types of graph-aware evaluation and governance protocols from a dynamic-graph perspective, which complement end-task evaluation. The goal is to provide a compact structural lens for designing and governing self-evolving agents.

\noindent\faGithub\ 
\textbf{Github:} 
\href{https://github.com/LuckyGirl-XU/Awesome-Agent-Dynamic-Graphs.git}
{\textcolor{myblue}{Self-Evolving-Agents-with-Dynamic-Graph}}
\end{abstract}


\IEEEpeerreviewmaketitle

\section{Introduction}

\IEEEPARstart{L}{arge} language model (LLM)-based agents are evolving from short-horizon task solvers into long-running systems that persist across sessions, accumulate experience, and adapt their behavior over time~\cite{xi2023agents,wang2024llmagentssurvey}. 
Their capabilities now include persistent memory, tool use, skill acquisition, workflow optimization, and multi-agent coordination~\cite{packer2023memgpt,park2023generative,hu2024aflow,wu2024autogen}. 
Consequently, agent states are both non-stationary and structurally evolving: stored knowledge, available capabilities, executable workflows, and inter-agent relations can all evolve through interaction, feedback, and environmental drift. Understanding such systems therefore requires a dynamic structural perspective that accounts for how agent states, capabilities, and relations change over time.

\begin{figure*}
    \centering
    \includegraphics[width=1\linewidth]{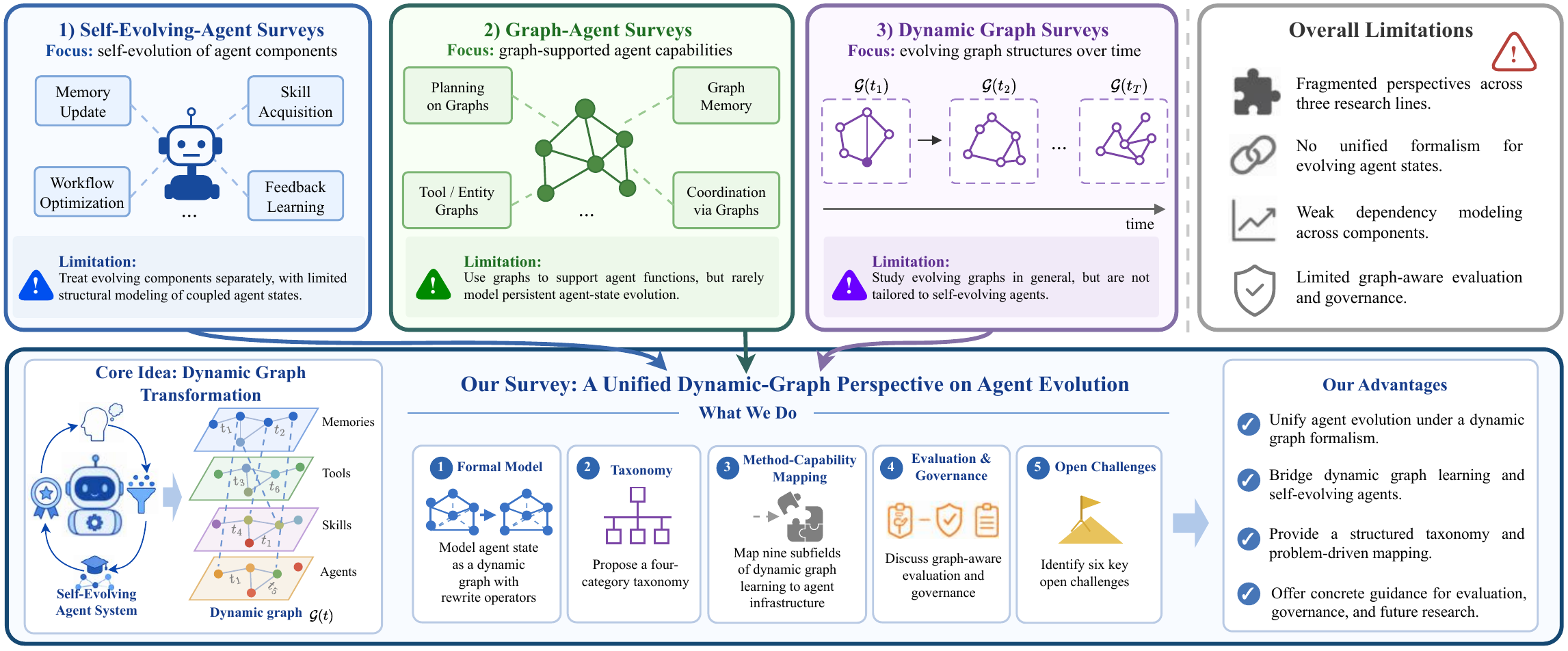}
    \caption{Positioning of this survey relative to existing LLM-agent, graph-agent, and dynamic-graph surveys.}
    \label{fig:intro_compare}
\end{figure*}

Existing surveys address this problem from two complementary but largely separate perspectives. Graph-agent surveys~\cite{bei2025graphsmeetagents,liu2025graphaugmentedagents} are mainly \emph{agent-functionality-centered}, studying how mostly static graphs support memory, retrieval, tool use, and multi-agent coordination. Self-evolving-agent surveys~\cite{gao2025selfevolvingagents,fang2025comprehensive} are mainly \emph{agent-evolution-mechanism-centered}, studying how agent components adapt via reflection, self-training, self-rewarding, and experience distillation.However, neither perspective fully captures a central property of self-evolving agents: persistent updates to memories, tools, skills, workflows, and inter-agent relations are structurally coupled and can propagate through dependencies over time. As a result, existing surveys provide limited support for modeling, analyzing, and governing agent systems whose states evolve as dynamic structures. This gap motivates a new perspective that jointly models agent functionality, state evolution, and dynamic graph topology, as illustrated in Fig.~\ref{fig:intro_compare}.

This survey aims to bridge this gap by framing \emph{agent evolution as dynamic graph transformation}. Thus, a self-evolving agent can be a structured system of coupled entities and relations. As these structures evolve via interaction, a \emph{dynamic graph perspective} becomes natural: typed nodes and edges encode heterogeneous states, temporal attributes track evolving evidence, and schema-constrained rewrites model activation, propagation, and rollback. This perspective gives rise to three motivations.

\noindent\textbf{Motivation \uppercase\expandafter{\romannumeral 1}: Unifying heterogeneous agent evolution.} 
Existing agent-evolution studies~\cite{shinn2023reflexion,zheng2024expel,hu2024aflow,liu2024dylan,wang2025agentdropout,lu2026dytopo,jiang2025gtd,fan2026todycomm,leong2025amas,wang2025unveiling,zhou2025guardian} span dynamic-graph-based agent methods, graph-structured agent modules, and non-graph mechanisms across memory, tool use, workflow optimization, communication topology, coordination efficiency, and safety governance. Yet these works remain fragmented, with different assumptions, objectives, and evaluation criteria. This motivates a dynamic-graph transformation lens under which heterogeneous evolution mechanisms can be studied together with their downstream dependencies, revisions, and rollbacks.

\noindent\textbf{Motivation \uppercase\expandafter{\romannumeral 2}: Connecting dynamic graph learning with self-evolving agents.} 
Dynamic graph learning~\cite{kazemi2020representation,qin2023temporal,ur2025primer,feng2025comprehensive,skarding2021foundations,rossi2020tgn,xu2020tgat,huang2023tgb,yu2023dygformer,xu2026unlocking} provides mature foundations for modeling evolving structures, including event streams, temporal dependencies, graph generation, and explanation. These foundations are directly relevant to self-evolving agents, whose memories, tools, skills, workflows, and inter-agent relations also change under new evidence, feedback, and drift~\cite{zhou2025guardian,nie2026skillgraph,li2026timem,rasmussen2025zep}. However, the two areas remain largely separate: dynamic graph learning rarely targets persistent agent-state evolution, while agent surveys rarely explain how dynamic graph methods can be reused as agent-evolution infrastructure. This motivates a systematic bridge that connects dynamic graph learning to the design, adaptation, diagnosis, and governance of self-evolving agents.

\noindent\textbf{Motivation \uppercase\expandafter{\romannumeral 3}: Toward structure-aware evaluation and governance.}
Current evaluation of self-evolving agents~\cite{liu2024agentbench,li2024agentboard,jimenez2024swebench,wu2024longmemeval} remains largely task- or answer-centered, which can miss failures in the evolving state: correct answers may rely on stale memories, unsafe updates may propagate through tools or workflows, and untraceable changes may persist across interactions. Recent safety works~\cite{ruan2024toolemu,debenedetti2024agentdojo,zhang2024asb} reinforce this need: tool-use benchmarks expose failures in invocation, stateful interaction, prompt injection, and harmful multi-step actions, while multi-agent studies~\cite{zhou2025guardian,he2025sentinelagent} show that errors and anomalies can propagate through communication structures. These findings require structure-aware evaluation and governance, where execution records are modeled as dynamic dependency graphs to support provenance tracing, affected-scope analysis, rollback, and audit~\cite{sequeira2026agent}.

\noindent\textbf{Contribution.} These motivations shape the organization of this survey. We first develop a dynamic-graph perspective on agent evolution, using dynamic graph transformation as a shared formal space for dynamic-graph-based, graph-based, and graph-transformable agent methods. This perspective represents evolving agent states through typed nodes, edges, subgraphs, and dependency-aware rewrites. It yields two core contributions: a four-pattern taxonomy of existing dynamic-graph-based agent works; and a systematic mapping from nine dynamic-graph-learning subfields to agent capabilities. We further extend this structural view to graph-aware evaluation and governance that complements end-task evaluation. Overall, dynamic graph transformation provides the organizing lens connecting agent-evolution mechanisms, dynamic graph learning, and governance protocols, as shown in Fig.~\ref{fig:intro_compare}.

This survey makes the following contributions.
\begin{itemize}
\item \textbf{Conceptual Framing:} We formalize agent evolution as dynamic graph transformation, representing evolving memories, tools, skills, workflows, and inter-agent relations as typed nodes, edges, and subgraphs.
\item \textbf{Systematic Taxonomy:} We organize existing self-evolving agent works through four graph-transformation patterns: node/feature evolution, edge/topology evolution, subgraph activation, and cross-component co-evolution.
\item \textbf{Infrastructure Bridge:} We connect dynamic graph learning with self-evolving agent infrastructure by mapping nine dynamic-graph-learning families to advance prediction, activation, generation, diagnosis, rollback, and governance capabilities.
\item \textbf{Evaluation and Governance:} We discuss five types of graph-aware evaluation and governance protocols for agent evolution: leakage-free temporal evaluation, privacy and deletion checking, safety monitoring, rollback analysis, and audit.
\item \textbf{Future Agenda:} We outline six open challenges for building reliable, auditable, and controllable self-evolving agents under the dynamic graph framing.
\end{itemize}

\noindent\textbf{Survey Scope.} This survey studies self-evolving agents from a dynamic-graph perspective, rather than providing a broad review of self-evolving agents~\cite{gao2025selfevolvingagents,fang2025comprehensive}, graph-supported agent systems~\cite{bei2025graphsmeetagents,liu2025graphaugmentedagents}, related agent subareas~\cite{yue2026workflow,yang2026graphmemory}, or general dynamic graph learning~\cite{kazemi2020representation,skarding2021foundations,qin2023temporal,ur2025primer,feng2025comprehensive}. 

\noindent\textbf{Organization.} Section~\ref{sec:prelim} introduces the background. Section~\ref{sec:agent-evolution} presents agent evolution as dynamic graph transformation, covering existing $46$ dynamic-graph-based agent works under a four-part taxonomy. Section~\ref{sec:dgl_infra} treats dynamic graph learning as agent-evolution infrastructure, organized by dynamic graph families. Section~\ref{sec:eval_gov} discusses graph-aware evaluation and governance, and Section~\ref{sec:conclusion} concludes.

\section{Preliminaries}
\label{sec:prelim}
\noindent\textbf{Dynamic Agent Graph.} 
We model the state representation of a self-evolving agent system at time $t$ as a dynamic graph
\begin{equation}
  \graphG(t) =
  \big(\nodeset(t), \edgeset(t), \boldsymbol{X}_V(t), \boldsymbol{X}_E(t), \boldsymbol{Y}(t)\big),
  \label{eq:agent-graph}
\end{equation}
where $\nodeset(t)$ and $\edgeset(t)$ are time-indexed nodes and edges, $\boldsymbol{X}_V(t)$ and $\boldsymbol{X}_E(t)$ denote node and edge attributes, and $\boldsymbol{Y}(t)$ denotes labels for nodes and edges. Eq.~\eqref{eq:agent-graph} can cover diverse dynamic graph types~\cite{huang2023tgb,rossi2020tgn,xu2025unidyg,you2022roland,wu2026understanding,zhang2024dtgb,huangsimple}, including continuous-time dynamic graphs (CTDGs), discrete-time dynamic graphs (DTDGs), dynamic heterogeneous graphs (DHGs), and dynamic text-attributed graphs (DyTAGs). 
With this dynamic graph view, we next map agent components to graph primitives. 
The node and edge schemas define the typed objects and relations in agent graphs, while the rewrite operators characterize how these structures evolve in the mechanisms discussed below.

\noindent\textbf{Graph-Conditioned Agent Policy.}
Given a user query or environmental input $q_t$, action history $h_t$, and activated support subgraph $\graphG_{\mathrm{act}}(t)$, the agent selects an action by
\begin{equation}
a_t \sim \mathcal{P}_\theta\!\left(
\cdot \mid q_t, h_t, \graphG_{\mathrm{act}}(t)
\right),
\label{eq:agent-policy}
\end{equation}
where $a_t$ may be a natural-language response, a tool call, a workflow step, or a message to another agent.
This makes the agent graph operational: graph state matters only insofar as it conditions future agent behavior.

\noindent\textbf{Node Schema.} 
A practical base schema for self-evolving agents includes four core node types: \texttt{memory}, \texttt{tool}, \texttt{skill}, and \texttt{agent}. 
A \texttt{memory} node represents stored knowledge that may be retrieved or revised over time.
A \texttt{tool} node describes an external callable resource, such as an API, database, or environment interface, together with its input--output schema, permissions, and execution history.
A \texttt{skill} node captures a reusable capability that can be invoked or adapted across tasks.
An \texttt{agent} node represents an autonomous or semi-autonomous actor with a role, policy, memory scope, tool access, and communication interface.
These nodes carry type-specific attributes, such as textual content, temporal validity, metadata, and prompts~\cite{rasmussen2025zep,zhang2024toolnet,wu2024autogen,liu2024dylan,zhou2025guardian,xu2025amem}.

\noindent\textbf{Edge Schema.}
Edges are typed relations among agent-state objects and may be directed when the relation is asymmetric or operationally ordered.
A compact edge schema can define three core edge families: \emph{dependency} edges for state, tool, skill, and workflow dependencies; \emph{communication} edges for inter-agent interaction; and \emph{provenance} edges for evidence tracing, updates, and rollback. Finer relations, such as reference and workflow control, can be modeled as subtypes. Edges may also carry features such as timestamps, validity intervals, activation status, or policy tags. When direction matters, edge orientation should follow a fixed operational convention.

\noindent\textbf{Rewrite Operators.} 
Evolution is modeled as the application of an ordered sequence of typed rewrites $\rho_1,\rho_2,\ldots$ to the agent graph.
A rewrite is a schema-constrained operation, such as node insertion/deletion, edge insertion/deletion, or feature update; dependency propagation is modeled as a cascade of such rewrites~\cite{heckel2006graph}. 
Subgraph activation is treated as a read-only graph operation that selects a relevant subgraph without committing a state change. 
The agent proposes candidate rewrites, and feedback from users, tools, or other agents determines whether they are accepted, rejected, or rolled back.

\noindent\textbf{Formal Semantics.}
We interpret the rewrite operators as typed graph-transformation rules under the double-pushout (DPO) formulation~\cite{taentzer2003agg,ehrig2006fundamentals}. 
A DPO rule is specified by a span of typed graph morphisms $L \leftarrow K \rightarrow R$, where $L$ is the pattern matched in $\graphG(t)$, $K$ is the preserved interface, and $R$ is the replacement structure. 
This formalism provides type safety for heterogeneous schemas and a basis for reasoning about concurrent rewrites through standard DPO concurrency conditions. 
In standard DPO graph transformation, the gluing conditions determine whether the pushout complement exists and hence whether a rule is applicable; in our agent setting, a violation is treated as an inadmissible rewrite proposal rejected by the runtime. 
Accepted rewrites are logged with their matches and affected subgraphs, providing graph-level evidence for audit and rollback. 
Throughout the survey, concrete update schemas are treated as DPO-style rules instantiated at specific matches. 
Let $\boldsymbol{s}_v(t)=(y_v(t),\boldsymbol{x}_v(t))$ and $\boldsymbol{s}_e(t)=(y_e(t),\boldsymbol{x}_e(t))$ denote node and edge states, where $y_v(t)$ and $y_e(t)$ are labels and $\boldsymbol{x}_v(t)$ and $\boldsymbol{x}_e(t)$ are attributes. 
For attributed graphs, $v[\boldsymbol{s}_v(t)]$ and $e[\boldsymbol{s}_e(t)]$ denote nodes and edges equipped with these internal states.

\noindent\textbf{Connection to Graph Databases.} 
The agent graph can be implemented as a temporal property graph using temporal graph-database models~\cite{debrouvier2021model}.
Graph transformation tools such as AGG~\cite{taentzer2003agg} provide executable DPO rule engines that can support implementation.

\begin{figure*}[t]
    \centering

\providecommand{\dynagent}[1]{#1}
\providecommand{\statagent}[1]{#1}

\resizebox{\textwidth}{!}{%
\begin{tikzpicture}[
    y=0.6cm,
    font=\scriptsize,
    every path/.style={line cap=round,line join=round},
    conn/.style={draw=black!65, line width=0.55pt},
    arr/.style={draw=black!65, line width=0.55pt, -{Stealth[length=2.0mm,width=1.4mm]}},
    rootbox/.style={
        rounded corners=4pt,
        draw=black!70,
        fill=gray!10,
        line width=0.65pt,
        minimum width=0.68cm,
        minimum height=9.5cm,
        align=center
    },
    catbase/.style={
        rounded corners=5pt,
        draw=black!62,
        line width=0.72pt,
        align=center,
        text width=4.75cm,
        minimum height=0.56cm,
        inner sep=1.00pt,
        font=\scriptsize\bfseries
    },
    subbase/.style={
        rounded corners=5pt,
        draw=black!62,
        line width=0.72pt,
        align=center,
        text width=5.75cm,
        minimum height=0.56cm,
        inner sep=0.75pt,
        font=\scriptsize
    },
    leafbase/.style={
        rounded corners=5pt,
        draw=black!62,
        line width=0.72pt,
        align=left,
        text width=6.95cm,
        minimum height=0.66cm,
        inner xsep=4.0pt,
        inner ysep=2.0pt,
        font=\scriptsize
    }
]

\def\xroot{0.0}
\def\xcat{3.35}
\def\xtrunk{6.35}
\def\xsub{9.95}
\def\xleaf{16.95}


\node[rootbox] (rootbox) at (\xroot,-8.20) {};
\node[rotate=90, align=center, font=\bfseries\scriptsize] at (rootbox.center)
{Agent Evolution as Dynamic Graph Transformation};

\node[catbase, fill=red!4, draw=red!55] (g1) at (\xcat,-2.70)
{A.1 Node \& Feature Evolution};

\node[subbase, fill=red!3, draw=red!55] (g1s1) at (\xsub,-0.70)
{Agent Memory Node};

\node[leafbase, fill=red!1, draw=red!55] (g1l1) at (\xleaf,-0.70)
{ \dynagent{TiMem~\cite{li2026timem}},
 \dynagent{Zep~\cite{rasmussen2025zep}},
 \dynagent{AriGraph~\cite{anokhin2024arigraph}},
 \dynagent{GAAMA~\cite{paul2026gaama}},
 \dynagent{GAM~\cite{wu2026gam}},
 \dynagent{GSEM~\cite{han2026gsem}},
 \dynagent{G-Memory~\cite{zhang2025gmemory}}
 };

\node[subbase, fill=red!3, draw=red!55] (g1s2) at (\xsub,-2.05)
{Agent Skill Node};

\node[leafbase, fill=red!1, draw=red!55] (g1l2) at (\xleaf,-2.05)
{\dynagent{SkillOps~\cite{pu2026skillops}},
 \dynagent{GoS~\cite{liu2026graph}}
 };

\node[subbase, fill=red!3, draw=red!55] (g1s3) at (\xsub,-3.40)
{Agent Tool Use Node};

\node[leafbase, fill=red!1, draw=red!55] (g1l3) at (\xleaf,-3.40)
{\dynagent{SEARL~\cite{feng2026searl}},
\dynagent{ToolNet~\cite{zhang2024toolnet}},
\dynagent{NaviAgent~\cite{jiang2025naviagent}}
};

\node[subbase, fill=red!3, draw=red!55] (g1s4) at (\xsub,-4.75)
{Multi-Agent Evolution};

\node[leafbase, fill=red!1, draw=red!55] (g1l4) at (\xleaf,-4.75)
{\dynagent{AgentDropout~\cite{wang2025agentdropout}},
\dynagent{DyLAN~\cite{liu2024dylan}}};

\node[catbase, fill=blue!4, draw=blue!55] (g2) at (\xcat,-8.15)
{A.2 Edge \& Topology Evolution};

\node[subbase, fill=blue!3, draw=blue!55] (g2s1) at (\xsub,-6.05)
{Workflow Optimization};

\node[leafbase, fill=blue!1, draw=blue!55] (g2l1) at (\xleaf,-6.05)
{\dynagent{AFlow~\cite{hu2024aflow}},
 \dynagent{DynTaskMAS~\cite{yu2025dyntaskmas}},
 \dynagent{AgentConductor~\cite{wang2026agentconductor}},
 \dynagent{GPTSwarm~\cite{zhuge2024gptswarm}},
 \dynagent{MASS~\cite{zhou2025mass}},
 \dynagent{ABSTRAL~\cite{song2026abstral}}
 };

\node[subbase, fill=blue!3, draw=blue!55] (g2s2) at (\xsub,-8.15)
{Multi-Agent Communication};

\node[leafbase, fill=blue!1, draw=blue!55] (g2l2) at (\xleaf,-8.15)
{\dynagent{DyLAN~\cite{liu2024dylan}},
 \dynagent{AgentPrune~\cite{zhang2024agentprune}},
 \dynagent{AgentDropout~\cite{wang2025agentdropout}},
 \dynagent{GTD~\cite{jiang2025gtd}},
 \dynagent{TodyComm~\cite{fan2026todycomm}},
 \dynagent{SafeSieve~\cite{zhang2026safesieve}},
 \dynagent{AgentNet~\cite{yang2026agentnet}},
 \dynagent{RUMAD~\cite{wang2026rumad}},
 \dynagent{ResMAS~\cite{zhou2026resmas}},
 \dynagent{DAGP~\cite{wang2025dagp}},
 \dynagent{AMAS~\cite{leong2025amas}},
 \dynagent{G-Designer~\cite{zhang2024gdesigner}},
 \dynagent{Graph-of-Agents~\cite{yun2026graphofagents}},
 \dynagent{AGP~\cite{li2025agp}},
 \dynagent{GoAgent~\cite{chen2026goagent}},
 \dynagent{TopoDIM~\cite{sun2026topodim}},
 \dynagent{HyperAgent~\cite{zhang2025hyperagent}}
 };

\node[subbase, fill=blue!3, draw=blue!55] (g2s3) at (\xsub,-10.35)
{Agent Skill/Tool Dependency};

\node[leafbase, fill=blue!1, draw=blue!55] (g2l3) at (\xleaf,-10.35)
{\dynagent{SkillOps~\cite{pu2026skillops}}, \dynagent{NaviAgent~\cite{jiang2025naviagent}}};

\node[catbase, fill=orange!5, draw=orange!70!black] (g3) at (\xcat,-12.38)
{A.3 Subgraph Activation};

\node[subbase, fill=orange!4, draw=orange!70!black] (g3s1) at (\xsub,-11.70)
{Evidence Activation: Agent Memory};

\node[leafbase, fill=orange!2, draw=orange!70!black] (g3l1) at (\xleaf,-11.70)
{\dynagent{ToG-3~\cite{wu2025tog3}},
\dynagent{GAM~\cite{wu2026gam}},
\dynagent{GAAMA~\cite{paul2026gaama}},
\dynagent{MemORAI~\cite{van2026memorai}},
 \dynagent{Zep~\cite{rasmussen2025zep}}};

\node[subbase, fill=orange!4, draw=orange!70!black] (g3s3) at (\xsub,-13.05)
{Execution Activation: Workflow/Skill/Team};

\node[leafbase, fill=orange!2, draw=orange!70!black] (g3l3) at (\xleaf,-13.05)
{\dynagent{DyLAN~\cite{liu2024dylan}},
 \dynagent{DyTopo~\cite{lu2026dytopo}},
 \dynagent{ARG-Designer~\cite{li2025argdesigner}}};

\node[catbase, fill=green!5, draw=green!55!black] (g4) at (\xcat,-15.08)
{A.4 Cross-Component Co-Evolution};

\node[subbase, fill=green!4, draw=green!55!black] (g4s1) at (\xsub,-14.40)
{Cross-Component Cascade Rewrites};

\node[leafbase, fill=green!2, draw=green!55!black] (g4l1) at (\xleaf,-14.40)
{\dynagent{MetaGen~\cite{wang2026metagen}},
 \dynagent{TacoMAS~\cite{xu2026tacomas}},
 \dynagent{SkillGraph~\cite{nie2026skillgraph}},
 \dynagent{EvoMAC~\cite{hu2024evomac}}};

\node[subbase, fill=green!4, draw=green!55!black] (g4s2) at (\xsub,-15.75)
{Safety-Triggered Propagation};

\node[leafbase, fill=green!2, draw=green!55!black] (g4l2) at (\xleaf,-15.75)
{\dynagent{GUARDIAN~\cite{zhou2025guardian}},
 \dynagent{SentinelAgent~\cite{he2025sentinelagent}},
 \dynagent{G-Safeguard~\cite{wang2025gsafeguard}}};

\draw[conn] (rootbox.east |- g1.west) -- (g1.west);
\draw[conn] (rootbox.east |- g2.west) -- (g2.west);
\draw[conn] (rootbox.east |- g3.west) -- (g3.west);
\draw[conn] (rootbox.east |- g4.west) -- (g4.west);


\draw[conn] (g1.east) -- (\xtrunk,-2.70);
\draw[conn] (\xtrunk,-0.70) -- (\xtrunk,-4.75);
\draw[conn] (\xtrunk,-0.70) -- (g1s1.west);
\draw[conn] (\xtrunk,-2.05) -- (g1s2.west);
\draw[conn] (\xtrunk,-3.40) -- (g1s3.west);
\draw[conn] (\xtrunk,-4.75) -- (g1s4.west);

\draw[conn] (g2.east) -- (\xtrunk,-8.15);
\draw[conn] (\xtrunk,-6.05) -- (\xtrunk,-10.35);
\draw[conn] (\xtrunk,-6.05) -- (g2s1.west);
\draw[conn] (\xtrunk,-8.15) -- (g2s2.west);
\draw[conn] (\xtrunk,-10.35) -- (g2s3.west);

\draw[conn] (g3.east) -- (\xtrunk,-12.38);
\draw[conn] (\xtrunk,-11.70) -- (\xtrunk,-13.05);
\draw[conn] (\xtrunk,-11.70) -- (g3s1.west);
\draw[conn] (\xtrunk,-13.05) -- (g3s3.west);

\draw[conn] (g4.east) -- (\xtrunk,-15.08);
\draw[conn] (\xtrunk,-14.40) -- (\xtrunk,-15.75);
\draw[conn] (\xtrunk,-14.40) -- (g4s1.west);
\draw[conn] (\xtrunk,-15.75) -- (g4s2.west);


\draw[arr] (g1s1.east) -- (g1l1.west);
\draw[arr] (g1s2.east) -- (g1l2.west);
\draw[arr] (g1s3.east) -- (g1l3.west);
\draw[arr] (g1s4.east) -- (g1l4.west);

\draw[arr] (g2s1.east) -- (g2l1.west);
\draw[arr] (g2s2.east) -- (g2l2.west);
\draw[arr] (g2s3.east) -- (g2l3.west);

\draw[arr] (g3s1.east) -- (g3l1.west);
\draw[arr] (g3s3.east) -- (g3l3.west);

\draw[arr] (g4s1.east) -- (g4l1.west);
\draw[arr] (g4s2.east) -- (g4l2.west);

\end{tikzpicture}%
}
    \caption{Method-level taxonomy of $46$ representative self-evolving-agent methods using dynamic topologies and graphs.  }
    \label{fig:chapter3-taxonomy-tree}
\end{figure*}
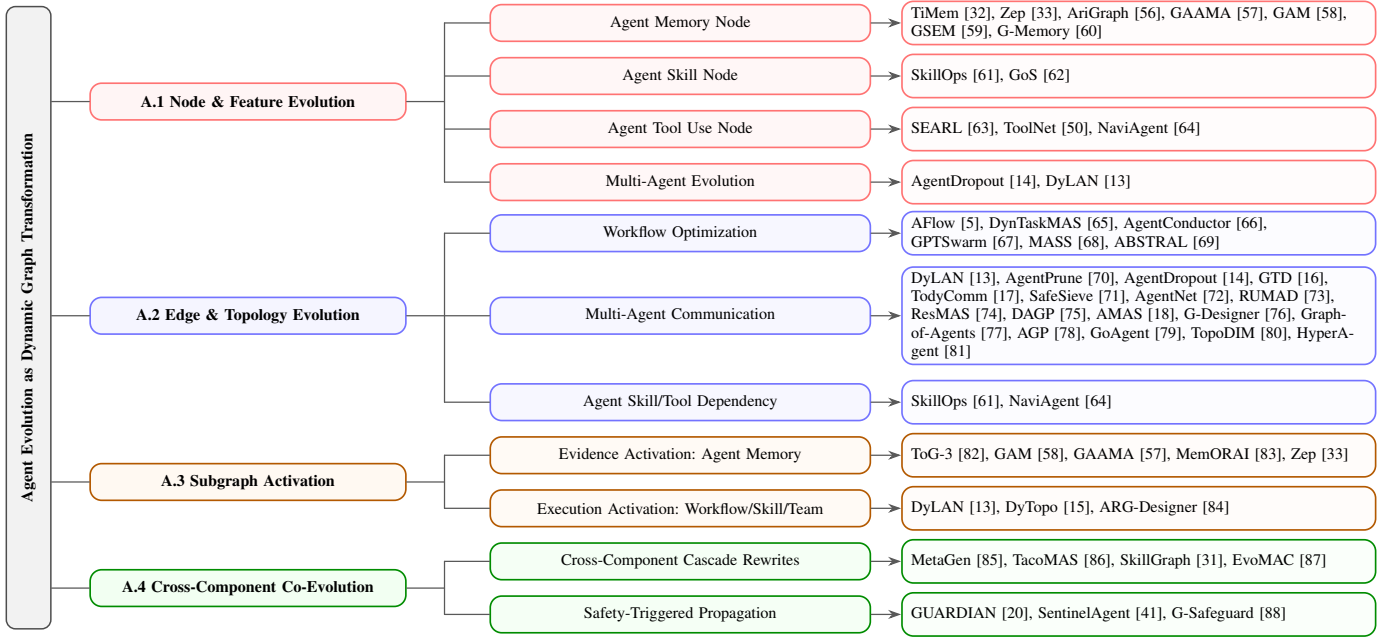

\section{Agent Evolution as Dynamic Graph Transformation}
\label{sec:agent-evolution} 
We organize existing self-evolving-agent methods based on dynamic graph/topology techniques by graph-transformation patterns: node/feature evolution, edge/topology evolution, subgraph activation, and cross-component co-evolution. Fig.~\ref{fig:chapter3-taxonomy-tree} groups $46$ representative methods into four branches, while Table~\ref{tab:rewrites} projects them into rewrite patterns.

\subsection{Node and Feature Evolution}
\label{sec:nodefeat}

Node and feature evolution, corresponding to Branch A.1 in Fig.~\ref{fig:chapter3-taxonomy-tree}, captures localized changes to individual agent-state components, represented as typed nodes and their attributes. It covers the insertion, revision, merging, or removal of memories, tools, skills, or agent states, along with the evidence needed to justify, validate, or roll back these updates. This abstraction unifies graph-native systems and persistent non-graph mechanisms by treating component updates as transformations over typed node states. Existing works~\cite{rasmussen2025zep,zhang2024toolnet,pu2026skillops,liu2024dylan} address such updates across memory, tool, skill, and agent adaptation, but often use component-specific terminology rather than an explicit graph-transformation view. We therefore introduce DPO-style schemas for insertion, deletion, feature update, and merge to characterize representative node-level self-evolving-agent works.

\noindent\textbf{DPO Rule Schemas for A.1.}
Let $\boldsymbol{s}_v(t)=(y_v(t),\boldsymbol{x}_v(t))$ denote the internal state of node $v$ at time $t$, where $y_v(t)$ is its label and $\boldsymbol{x}_v(t)$ is its attribute vector. 
At the node level, the main typed DPO schemas are:
\begin{align}
&{\scriptsize\textsc{Insert}}_{y,\boldsymbol{x}}:
L=K=\emptyset,
R=\{v_\mathrm{new}[(y,\boldsymbol{x})]\},\label{eq:a1-insert}\\
&{\scriptsize\textsc{Delete}}_{v}:
L=\{v[\boldsymbol{s}_v(t)]\}, K=R=\emptyset,\label{eq:a1-delete}  \\
&{\scriptsize\textsc{FeatureUpdate}}_{v,\phi}:
L=\{v[(y_v(t),\boldsymbol{x}_v(t))]\}, K=\{v\},\notag\\
&\qquad\qquad\quad R=\{v[(y_v(t),\phi(\boldsymbol{x}_v(t)))]\},\label{eq:a1-fupd}\\
&{\scriptsize\textsc{Merge}}_{v_1,v_2,\psi}:
L=\{v_1[\boldsymbol{s}_{v_1}(t)],v_2[\boldsymbol{s}_{v_2}(t)]\},
K=\{v_1,v_2\},\notag\\
&
R=\{v_1[\mathrm{arch}(\boldsymbol{s}_{v_1}(t))],
v_2[\mathrm{arch}(\boldsymbol{s}_{v_2}(t))],
v^*[\psi(\boldsymbol{s}_{v_1}(t),\boldsymbol{s}_{v_2}(t))]\}\notag\\
&\qquad\qquad\qquad
\cup\{e^{\mathrm{prov}}_{*i}: v^*\!\to\!v_i\}_{i=1,2}.
\label{eq:a1-merge}
\end{align}
Here, $L$, $K$, and $R$ are the matched pattern, preserved interface, and replacement graph. 
$v_\mathrm{new}$ is initialized by $(y,\boldsymbol{x})$, $\phi$ updates attributes, and $\psi$ consolidates matched states into $v^*$. 
$\mathrm{arch}(\cdot)$ archives the source node by marking it as inactive while preserving its identity and label, 
and $e^{\mathrm{prov}}_{*i}$ records provenance. 
Evidence-grounded insertion uses a contextual rule with non-empty $L=K$ or a later \textsc{Link}. 
By the DPO dangling condition, \textsc{Delete} applies only to nodes with no incident edges unless those edges are also removed. 
\textsc{Merge} is archival consolidation rather than node identification; dependency redirection is handled by later edge rewrites or cascades. 
We omit \textsc{Split}, since archived states and provenance already support undoing mistaken merges. We next discuss four node-level cases: agent memory, agent skill, agent tool use, and multi-agent evolution.

\noindent\textbf{Agent Memory.}
Agent memory is an important setting for node-level self-evolution. 
Existing graph-based memory works can be broadly categorized into relation-centric and consolidation-centric memory graphs.
The former~\cite{rasmussen2025zep,anokhin2024arigraph} links memories to entities, events, tasks, and provenance relations for time-aware recall and update tracking; for example, Zep~\cite{rasmussen2025zep} builds an evolving temporal knowledge graph over memory events, entities, and provenance.
The latter~\cite{li2026timem,paul2026gaama,wu2026gam,han2026gsem,zhang2025gmemory} compresses or organizes low-level observations into higher-level structures for long-term recall and reuse; for example, TiMem~\cite{li2026timem} organizes interaction histories into time-aware memory levels for temporal-hierarchical consolidation.
These graph-based systems make the underlying node-level operations explicit: new observations instantiate \textsc{Insert}, consolidation or abstraction instantiates \textsc{Merge}, and correction, decay, quality revision, or invalidation instantiates \textsc{FeatureUpdate} or \textsc{Delete}. 
Non-graph memory systems~\cite{packer2023memgpt,kang2025memory,lin2026memma,zhang2026memrl} can be mapped similarly by treating persistent memory items as typed nodes with evolving attributes. 
Recent benchmarks~\cite{maharana2024locomo,wu2024longmemeval,wu2024membench} further show growing attention to long-term memory update behavior.

\noindent\textbf{Agent Skill.}
Agent skills can be represented as capability nodes whose states encode scope, contract, prerequisites, and validation status.
Existing skill-related works can be approximately grouped into graph-structured skill retrieval/maintenance and experience-driven skill acquisition.
The first line~\cite{pu2026skillops,liu2026graph} organizes reusable skills and dependencies; the second line~\cite{wang2024voyager,zheng2024expel,zhang2026coevoskills,alzubi2026evoskill,xia2026skillrl,kuroki2025agent} learns or revises skills from interaction experience and can be mapped by treating acquired skills as typed nodes with evolving attributes.
Under the node-level schemas in Eqs.~\eqref{eq:a1-insert}-\eqref{eq:a1-merge}, new, revised, and consolidated skills correspond to \textsc{Insert}, \textsc{FeatureUpdate}, and \textsc{Merge}, respectively.
This framing records skill provenance and dependencies, enabling targeted revalidation when related memories, tools, or workflows evolve.

\noindent\textbf{Agent Tool Use.}
Agent tool use is a node-level case of self-evolution, where each tool can be treated as a state object encoding its interface, schema, constraints, and reliability.
Existing tool-use works can be classified into graph-native tool organization and API-grounded tool learning.
Graph-native methods~\cite{zhang2024toolnet,feng2026searl,jiang2025naviagent} organize tools, toolchains, or navigation relations as updatable graph structures. Concretely, SEARL~\cite{feng2026searl} maintains tool graph memory so execution feedback can update tool-use structure for later decisions.
Additionally, API-grounded methods~\cite{schick2023toolformer,patil2023gorilla,qin2024toolllm} improve tool selection or API use without maintaining an explicit evolving tool graph, but can be mapped by treating tool profiles and usage records as typed nodes.
Under our schemas, API or documentation drift instantiates \textsc{FeatureUpdate}, reliability revision updates tool attributes, and obsolete tools are deactivated or removed.
Tool-use traces and function-call benchmarks~\cite{li2023apibank,lu2024toolsandbox,yan2024bfcl,xiu2026astra} provide evidence for validating such updates.
This framing treats tool evolution as a persistent state change that may trigger revalidation of dependent skills and workflows (Section~\ref{sec:coevolve}) .

\noindent\textbf{Multi-Agent Evolution.}
Agents can be modeled as persistent nodes whose states encode role, expertise, permissions, and participation status.
Existing multi-agent works fall into dynamic agent participation and protocol-level collaboration.
The former~\cite{liu2024dylan,wang2025agentdropout} change which agents participate, specialize, or remain active, where DyLAN~\cite{liu2024dylan} builds a dynamic agent network whose active agents and communication relations adapt during collaboration.
Furthermore, protocol-level frameworks~\cite{liu2026learning,gao2024agentscope,li2026r,yu2025table,zhang2026verified} support collaboration policies such as decentralized coordination, procedural orchestration, or plan--execute--verify--replan control.
Under our dynamic-graph view, these mechanisms can be represented as updates to agent-node attributes and to communication, dependency, or provenance edges.
This connects node-level agent revision to the topology evolution (Section~\ref{sec:edgetopo}).

\begin{insightbox}
\textbf{Open Gaps.}
Node-level evolution exposes a key gap in self-evolving agents: existing systems can update memories, tools, skills, or agent states, but rarely attach calibrated validity, evidence strength, or affected-scope information to these updates.
Thus, local edits may silently affect dependent skills, workflows, or team-selection policies without triggering revalidation.
The open problem is affected-scope estimation: after a node update, the agent should identify which downstream components require rechecking, repair, or rollback, without turning each local edit into global revalidation.
Dynamic graph methods provide a natural basis by modeling updates as time-stamped events and propagating their influence via dependency and provenance structures.
\end{insightbox}

\subsection{Edge and Topology Evolution}
\label{sec:edgetopo}

Edge and topology evolution, corresponding to Branch A.2 in Fig.~\ref{fig:chapter3-taxonomy-tree}, covers self-evolving agent problems where persistent change lies in the relations among components rather than in individual component states.
Such evolution reorganizes memory support, tool--skill dependencies, workflow routing, and inter-agent coordination over time.
Compared with node and feature evolution, the focus shifts from local state revision to dependency, communication, and execution topology.
Existing work addresses these changes in workflow optimization, multi-agent coordination, and dependency management~\cite{hu2024aflow,yu2025dyntaskmas,zhang2024agentprune,pu2026skillops,jiang2025naviagent,jiang2025gtd,agentasagraph2025,graphplanner2026}.
We therefore introduce DPO-style schemas for edge-level transformations and use them to characterize existing topology-level self-evolving agent works.

\noindent\textbf{DPO Rule Schemas for A.2.}
Let $e_{ij}[\boldsymbol{s}_{ij}(t)]$ denote an edge between $v_i$ and $v_j$, where $\boldsymbol{s}_{ij}(t)=(y_{ij}(t),\boldsymbol{x}_{ij}(t))$ is its edge state, consisting of an edge label and edge attributes. At the edge level, we define four typed schemas:
\begin{align}
&{\scriptsize\textsc{Link}}_{y,\boldsymbol{x}}:
L=K=\{v_i,v_j\},
R=K\cup\{e_{ij}[(y,\boldsymbol{x})]\},
\label{eq:a2-link}\\
&{\scriptsize\textsc{Unlink}}_{e_{ij}}:
L=\{v_i,v_j,e_{ij}[\boldsymbol{s}_{ij}(t)]\},\notag\\
&\qquad\qquad K=\{v_i,v_j\},
R=K,
\label{eq:a2-unlink}\\
&{\scriptsize\textsc{Rewire}}_{e_{ij}\to e'_{ik}}:
L=\{v_i,v_j,v_k,e_{ij}[\boldsymbol{s}_{ij}(t)]\},\notag\\
&\qquad\qquad
K=\{v_i,v_j,v_k\},
R=K\cup\{e'_{ik}[\boldsymbol{s}_{ij}(t)]\},
\label{eq:a2-rewire}\\
&{\scriptsize\textsc{EdgeFeatureUpdate}}_{e_{ij},\phi}:
L=\{e_{ij}[(y_{ij}(t),\boldsymbol{x}_{ij}(t))]\},\notag\\
&\qquad\qquad
K=\{e_{ij}\}, R=\{e_{ij}[(y_{ij}(t),\phi(\boldsymbol{x}_{ij}(t)))]\}.
\label{eq:a2-eupd}
\end{align}
Here, \textsc{Link} adds a typed edge, \textsc{Unlink} removes an edge while preserving its endpoints, \textsc{Rewire} replaces $e_{ij}$ with $e'_{ik}$, and \textsc{EdgeFeatureUpdate} revises edge attributes while preserving edge identity.
Edge direction follows the semantics of the relation, e.g., sender-to-receiver.
The key constraint is not edge removal itself, but downstream consistency: removed or redirected edges must not remain referenced by provenance or dependency records.
Otherwise, the edit becomes a cross-component cascade (Section~\ref{sec:coevolve}).

\noindent\textbf{Workflow Optimization.}
Workflow optimization covers self-evolving-agent methods that revise executable structures rather than only selecting a fixed plan at inference time.
Existing works can be roughly grouped into workflow/architecture search and task-graph orchestration.
The former~\cite{hu2024aflow,zhuge2024gptswarm,zhou2025mass,song2026abstral,hu2024adas,shang2024agentsquare} searches over agentic workflows, computational graphs, or modular agent designs, while the latter~\cite{yu2025dyntaskmas,wang2026agentconductor} constructs task-dependent execution structures or role/task graphs for multi-agent coordination.
For example, DynTaskMAS~\cite{yu2025dyntaskmas} uses a dynamic task graph to decompose tasks, preserve dependencies, and support asynchronous parallel execution.
Under our schema, workflow optimization becomes edge rewriting over an executable task graph: components are selected, ordered, connected, or revised while workflow variants remain comparable and dependency changes traceable.

\noindent\textbf{Multi-agent Communication.}
Multi-agent communication evolution concerns how agents revise information exchange over time.
Existing works span from constructing task-adaptive communication structures to pruning or sparsifying inefficient ones.
For construction and routing methods~\cite{liu2024dylan,jiang2025gtd,fan2026todycomm,yang2026agentnet,wang2026rumad,zhou2026resmas,leong2025amas,zhang2024gdesigner,yun2026graphofagents,chen2026goagent,sun2026topodim,zhang2025hyperagent}, they generate, route, or adapt communication structures according to task context, agent roles, or robustness objectives.
For example, GTD~\cite{jiang2025gtd} formulates multi-agent topology synthesis as conditional graph generation, producing task-adaptive communication graphs.
For pruning and sparsification methods~\cite{zhang2024agentprune,wang2025agentdropout,zhang2026safesieve,wang2025dagp,li2025agp}, they remove redundant, costly, unsafe, or low-utility agents or links.
Under our schema, these methods instantiate \textsc{Link}, \textsc{Unlink}, \textsc{Rewire}, or \textsc{EdgeFeatureUpdate} over agent--agent communication edges, distinguishing persistent communication-topology evolution from workflow rewriting and temporary team activation.

\noindent\textbf{Skill--Tool Dependencies.}
Skill--tool dependencies focus on how self-evolving agents maintain relations among skills, tools, preconditions, and data requirements.
This includes (1) skill-side dependency maintenance~\cite{pu2026skillops}, (2) tool-side dependency maintenance~\cite{jiang2025naviagent}, and (3) tool--agent dependency retrieval~\cite{agentasagraph2025}, where skills, tools, and parent agents are linked through prerequisites, compatibility, invocation, or ownership relations.
SkillOps~\cite{pu2026skillops} represents skills as typed contracts in a hierarchical skill ecosystem graph for library-time maintenance, while NaviAgent~\cite{jiang2025naviagent} maintains a tool navigation structure that updates toolchain selection from execution feedback.
Under our schema, newly discovered dependencies instantiate \textsc{Link}, invalid dependencies instantiate \textsc{Unlink}, alternative toolchains instantiate \textsc{Rewire}, and reliability changes instantiate \textsc{EdgeFeatureUpdate}.
This framing supports targeted revalidation when a skill is revised, a tool drifts, or a dependency is rerouted.

\begin{insightbox}
\textbf{Open Gaps.}
Edge/topology evolution exposes a gap between topology editing and downstream validation.
Existing methods can construct, route, prune, or reweight workflow and communication edges, but rarely assess whether an edited edge is redundant, safety-critical, or part of a necessary reasoning or execution path.
Such edits may therefore break dependent skills, workflows, provenance chains, or coordination patterns without detection.
The open challenge is impact-aware, reversible topology maintenance: modeling topology edits as temporal events, propagating their effects through dependency and provenance structures, and maintaining versioned edge histories for monitoring and rollback.
\end{insightbox}

\begin{takeawaybox}
 \noindent\textbf{Takeaway.}
Although the methods in Section~\ref{sec:edgetopo} are presented as workflow optimization, communication routing or pruning, and skill--tool dependency maintenance, they share an evolving-topology view. 
Their edges represent execution, communication, or dependency relations, and their evolution corresponds to edge insertion, deletion, rewiring, or reweighting under the schemas in Eqs.~\eqref{eq:a2-link}-\eqref{eq:a2-eupd}.
Viewing these mechanisms as dynamic graph transformations makes structural changes explicit, helping identify affected components, check dependency consistency, and trigger targeted revalidation after topology updates.
\end{takeawaybox}

\begin{table*}[!t]
\centering
\caption{Representative dynamic or time-aware agent-evolution mechanisms as graph rewrites. 
The \emph{Scope} column uses graph-level terms to denote the affected agent-state subgraph: memory, skill, workflow, communication, agent, or trace graph.}
\label{tab:rewrites}
\footnotesize
\setlength{\tabcolsep}{3pt}
\renewcommand{\arraystretch}{1.12}

\begin{tabularx}{\textwidth}{
p{0.12\textwidth}
p{0.12\textwidth}
p{0.22\textwidth}
p{0.15\textwidth}
p{0.08\textwidth}
X}
\toprule
\textbf{Mechanism} &
\textbf{Trigger} &
\textbf{Rewrite} &
\textbf{Scope} &
\textbf{Persistence} &
\textbf{Examples} \\
\midrule

\rowcolor{gray!12}
\multicolumn{6}{l}{\textbf{A.1 Node and feature evolution}} \\

Memory update &
New evidence &
\textsc{Insert}/\textsc{Merge}/\textsc{FeatureUpdate} &
Memory graph &
Long &
Zep~\cite{rasmussen2025zep}, TiMem~\cite{li2026timem} \\

Skill update &
Trajectory/feedback &
\textsc{Insert}; \textsc{FeatureUpdate} &
Skill graph &
Long &
SkillOps~\cite{pu2026skillops} \\

\addlinespace[1pt]
\rowcolor{gray!12}
\multicolumn{6}{l}{\textbf{A.2 Edge and topology evolution}} \\

Workflow rewrite &
Task change &
\textsc{Link}/\textsc{Unlink}/\textsc{Rewire} &
Workflow graph &
Medium &
AFlow~\cite{hu2024aflow}, DynTaskMAS~\cite{yu2025dyntaskmas} \\

Communication pruning &
Redundant messages &
\textsc{Unlink}/\textsc{Rewire} &
Communication graph &
Medium &
AgentPrune~\cite{zhang2024agentprune}, AgentDropout~\cite{wang2025agentdropout} \\

Topology routing &
Round context &
\textsc{Link}/\textsc{Rewire} &
Communication graph &
Medium &
 GTD~\cite{jiang2025gtd} \\

\addlinespace[1pt]
\rowcolor{gray!12}
\multicolumn{6}{l}{\textbf{A.3 Read-only subgraph activation}} \\

Team activation &
Query/round &
\textsc{Activate} &
Agent graph &
Temporary &
DyLAN~\cite{liu2024dylan}, DyTopo~\cite{lu2026dytopo} \\

\addlinespace[1pt]
\rowcolor{gray!12}
\multicolumn{6}{l}{\textbf{A.4 Cross-component co-evolution}} \\

Workflow$\to$team &
Expertise change &
\textsc{Rewire}+cascade &
Workflow-Agent graph &
Medium &
MetaGen~\cite{wang2026metagen}, TacoMAS~\cite{xu2026tacomas} \\

Safety propagation &
Unsafe trace &
\textsc{FeatureUpdate} + cascade &
Trace graph &
Medium &
GUARDIAN~\cite{zhou2025guardian}, SentinelAgent~\cite{he2025sentinelagent} \\

\bottomrule
\end{tabularx}
\end{table*}

\subsection{Subgraph Activation}
\label{sec:subgraph}
Subgraph activation (Branch A.3 in Fig.~\ref{fig:chapter3-taxonomy-tree}) captures task-time adaptation in which an agent system selects a transient support subgraph from a persistent graph for reasoning or execution.
Here, the central problem of subgraph activation is relevance-conditioned selection: which subset of available states, relations, and components should be activated for the current task, and how this temporary context should constrain the next decision.
Existing systems~\cite{wu2025tog3,rasmussen2025zep,liu2024dylan,lu2026dytopo,sun2026topodim} often implement this step through retrieval, planning, routing, or coordination procedures, but rarely formulate it as an explicit graph operation.
We therefore formalize subgraph activation as a read-only selection schema and use it to characterize representative task-time activation mechanisms.

\noindent\textbf{Read-only Activation Schema for A.3.}
Read-only activation describes task-time support-subgraph selection without committing any structural change.
For a task $q$ and activation policy $\pi$, the agent selects a support subgraph $\graphG_{\mathrm{act}}(t)\subseteq\graphG(t)$.
In DPO-style notation, this can be represented as an identity production over the selected pattern:
\begin{equation}
{\scriptsize\textsc{Activate}}:\;
L_{\pi}=K_{\pi}=R_{\pi},
\;
\mathrm{img}(\mu_{q,\pi})=\graphG_{\mathrm{act}}(t),
\label{eq:a3-activation}
\end{equation}
where $\mu_{q,\pi}:L_{\pi}\to \graphG(t)$ is the match chosen by policy $\pi$ for task $q$.
Thus, $q$ and $\pi$ determine the match, not a persistent rewrite.
Its advantage is auditability: evidence use, tool routing, and team selection can be inspected without being mistaken for persistent learning.


\noindent\textbf{Activation Pattern.}
Most A.3 methods can be viewed as selecting anchors and then exposing limited context around them.
An activation policy $\pi$ first scores candidate memories, tools, skills, or agents for the task $q$, and then expands the selected anchors with relevant local structure to form $\graphG_{\mathrm{act}}(t)$.
This pattern explains how retrieval, workflow-fragment selection, and communication-neighborhood activation can share the same read-only graph interface~\cite{wu2025tog3,lu2026dytopo}.
The key distinction is that only the activated view changes; the persistent graph remains unchanged. Next, we discuss two cases: memory evidence activation and execution activation.

\noindent\textbf{Memory Evidence Activation.}
Memory evidence activation selects task-relevant memory states and relations without modifying the persistent memory store.
Existing works can be classified into graph-based evidence expansion and memory-store retrieval.
The first line~\cite{wu2025tog3,rasmussen2025zep} activate relational support structures from persistent or per-query graphs; for example, ToG-3~\cite{wu2025tog3} iteratively refines a query-specific support graph through multi-agent context retrieval and expansion.
We treat such methods as A.3 only when the constructed graph serves as transient evidence for the current agent decision rather than a committed memory update.
For memory-store methods~\cite{packer2023memgpt,zhong2023memorybank,kang2025memory}, they retrieve persistent memory items without necessarily exposing an explicit graph, but can be mapped by treating retrieved chunks as temporarily activated nodes.
Under our activation schema in Eq.~\eqref{eq:a3-activation}, the selected evidence is recorded as $\graphG_{\mathrm{act}}(t)$, making missing evidence, stale memories, or overly broad retrieval diagnosable.

\noindent\textbf{Execution Activation: Workflow/Skill/Team.}
Execution activation chooses task-specific components, e.g., a tool, without committing a persistent state change.
Existing works may be divided into graph-native execution activation and component-library activation.
For the first line~\cite{liu2024dylan,lu2026dytopo,li2025argdesigner}, they activate task-specific agents or communication structures. Here, ARG-Designer~\cite{li2025argdesigner} constructs a query-conditioned collaboration graph, which we treat as A.3 when the graph is used as a task-time execution view rather than stored as a persistent topology.
Second-line works~\cite{wang2024voyager,wang2026dyflow,zhang2026vipact} typically select executable skills, modules, or workflow steps. Taking DyFlow~\cite{wang2026dyflow} as an example, it provides a workflow-oriented case where selected execution flows can be viewed as activated nodes and edges for the current run.
These remain A.3 operations when the selected structure is only used at task time; once written back as a skill, workflow edge, or communication link, the update belongs to A.1, A.2, or A.4.

\begin{takeawaybox}
\noindent\textbf{Takeaway.}
Subgraph activation highlights an evaluation gap in self-evolving agents: existing benchmarks~\cite{liu2024agentbench,mialon2024gaia,jimenez2024swebench,yao2024taubench,xie2024osworld,deng2023mind2web,qiao2025benchmarking} mostly measure final task success, but rarely test whether the agent activated the right memories, tools, skills, agents, or workflow fragments.
A graph-based formulation makes this intermediate decision measurable via support-subgraph precision, required-evidence coverage, minimality, and temporal validity.
Dynamic graph methods~\cite{li2025ranking,chen2024rush} may provide useful signals, but the core challenge is designing activation-specific metrics for self-evolving agents.
\end{takeawaybox}

\subsection{Cross-Component Co-Evolution}
\label{sec:coevolve}

In self-evolving agents, a local update is often not self-contained: revising a memory may affect skills that rely on it, changing a tool may invalidate workflows that call it, and modifying an agent role may require communication or team structures to be updated. Therefore, the last branch in Fig.~\ref{fig:chapter3-taxonomy-tree} captures persistent system-level evolution triggered by such cross-component dependencies.
The central problem is propagation control: after an initial update, the agent must identify which downstream components (e.g., workflows) should be revised, revalidated, or rolled back to preserve consistency and safety.
Existing methods~\cite{wang2026metagen,xu2026tacomas,nie2026skillgraph,hu2024evomac,wang2026topoevo,zhou2025guardian,he2025sentinelagent,wang2025gsafeguard,madaan2023selfrefine,liu2026synthesizing} implement related effects through reflection, feedback, optimization, or monitoring, but rarely expose the dependency paths through which these updates propagate as explicit graph transformations.
We therefore model cross-component co-evolution as typed cascades of DPO rewrites, providing a graph-level view of system-level self-evolution.

\noindent\textbf{Cascade Rewrite Schema for A.4.}
A cross-component cascade is a timestamped trace of DPO rewrites triggered by an initial cause $c$:
\begin{equation}
\mathcal{C}(c)=\bigl((\rho_i,\mu_i,t_i)\bigr)_{i=1}^{k},
\;
\mu_i:L_i\to\graphG(t_i).
\label{eq:a4-cascade}
\end{equation}
Here $\rho_i$ is the $i$-th rewrite, $\mu_i$ is its match, and $\graphG(t_i)$ denotes the host graph state immediately before applying $\rho_i$.
The affected subgraph is $S_i=\mathrm{img}(\mu_i)$.
Let $\tau(S_i)$ denote the set of component types appearing in $S_i$, such as memory, tool, skill, workflow, or agent.
A trace belongs to A.4 when it spans multiple component types,
$\left|\bigcup_{i=1}^{k}\tau(S_i)\right|\ge 2$,
and the sequence is not a set of independent edits: at least one later rewrite depends on, or conflicts with, an earlier rewrite.
Such dependency or conflict can be identified from dependency, composition, provenance, invocation, or explicit cause records.
This schema ties revalidation and rollback targets to explicit affected subgraphs.

\noindent\textbf{Cross-Component Cascade Rewrites.}
Cross-component cascade rewrites capture updates whose effects propagate beyond the initially edited component.
Existing works generally fall into feedback-to-capability propagation and capability/topology co-adaptation.
The first line~\cite{nie2026skillgraph} turns task feedback, failures, or trajectories into reusable skills, rules, or agent expertise, while the second line~\cite{xu2026tacomas,wang2026metagen,hu2024evomac} uses updated capability, utility, or topology signals to revise roles, communication links, or collaboration structures.
For example, TacoMAS~\cite{xu2026tacomas} couples a fast expertise-update loop with a slower topology-editing loop, making capability changes and communication-structure updates part of the same evolving graph process.
Other feedback-driven self-improvement methods~\cite{li2024agenthospital,qiao2024autoact} can be mapped similarly when feedback-derived artifacts are materialized and reused to update dependent memories, skills, workflows, or evaluation components.
Under our schema in Eq.~\eqref{eq:a4-cascade}, these methods are unified as cascade traces $\mathcal{C}(c)$ linking the triggering cause, affected subgraphs, and revalidation or rollback targets.

\noindent\textbf{Safety-Triggered Propagation.}
Safety-triggered cascades arise when unsafe tool use, API drift, or abnormal interaction invalidates dependent skills, workflows, or policies.
Existing dynamic-graph-based methods~\cite{zhou2025guardian,he2025sentinelagent,wang2025gsafeguard} model unsafe behavior over temporal interaction or execution graphs. We take GUARDIAN~\cite{zhou2025guardian} as an example: it models multi-agent collaboration as a temporal attributed graph to detect hallucination and error propagation, providing anomaly signals for later review or mitigation.
Additionally, there are non-graph studies~\cite{ghosh2025safety,yin2026policy} that address related risks through attack synthesis, policy design, evaluation, or mitigation, but do not explicitly expose graph-state propagation.
Under our schema, an anomaly becomes the cause $c$, and dependency or provenance edges determine which downstream components require revalidation or possible rollback.
This graph view identifies the affected subgraph, enabling localized repair instead of global retraining or broad manual inspection.

\begin{insightbox}
\textbf{Open Gaps.}
A.4 methods show cross-component adaptation, but rarely expose the actual cascade trace $\mathcal{C}(c)$ that connects a trigger to its downstream edits~\cite{zhou2025guardian,he2025sentinelagent}.
For example, if a tool contract changes, an agent should not only mark the tool as outdated, but also identify the skills that call it, the workflows that contain those skills, and the teams that depend on those workflows.
The open problem is affected-scope estimation for revalidation: after a local update, the agent should check enough downstream components to prevent silent regression, while avoiding unnecessary global repair.
\end{insightbox}

\subsection{Putting the Four Rewrite Branches Together}
\label{sec:worked}
The four branches above can be read as a progression from local state change to relation updates, read-only activation, and cross-component propagation.
As summarized in Table~\ref{tab:rewrites}, A.1 covers persistent component-state updates: new evidence or trajectory feedback triggers memory or skill \textsc{Insert}, \textsc{Merge}, or \textsc{FeatureUpdate} operations over component graphs.
A.2 moves from states to relations: task changes, redundant messages, or round-level context trigger workflow rewrites, communication pruning, or topology routing through \textsc{Link}, \textsc{Unlink}, and \textsc{Rewire}.
A.3 differs from both because it is temporary: a query or execution round activates a task-specific support subgraph, such as relevant memories, tools, workflow fragments, or a temporary team, without committing a persistent update.
A.4 captures cascades where an initial change propagates across at least two component types, such as expertise changes that affect team or workflow organization, or unsafe traces that trigger revalidation across dependent skills, workflows, or policies.
Table~\ref{tab:rewrites} therefore highlights four dimensions that distinguish the branches: trigger, rewrite operator, affected scope, and persistence or propagation pattern.
A full agent system may involve all four branches: it may activate support for one task, update memories or skills after feedback, rewire workflows or communication links, and trigger a cascade when those changes affect dependent components.
Thus, A.1--A.4 are not mutually exclusive system labels; they identify the primary graph-rewrite role of each mechanism, while a full agent system may span several branches.

\begin{takeawaybox}
\textbf{Takeaway.}
A.1--A.4 provide a graph-rewrite view of self-evolving agents: maintaining component states, adapting relations, activating task-time support, and controlling cross-component cascades.
Together, they clarify what changes, how it changes, whether the change is persistent, and how its effects propagate.
Table~\ref{tab:rewrites} makes this view concrete by mapping representative mechanisms to their triggers, rewrite operators, affected scopes, and persistence levels.
This taxonomy gives a common language for comparing existing dynamic-graph-based agent methods and identifying where validation, rollback, and governance should be applied.
\end{takeawaybox}

\section{Dynamic Graph Learning as Agent Infrastructure}
\label{sec:dgl_infra}

Section~\ref{sec:agent-evolution} defines a graph-transformation interface for self-evolving agents, exposing control points such as support activation, dependency rewiring, cascade scoping, and rollback.
In this section, we build a capability mapping from dynamic graph learning (DGL) to these control points, using DGL as reusable infrastructure for self-evolving agents.
Following dynamic-graph surveys and benchmarks~\cite{kazemi2020representation,skarding2021foundations,longa2023graphlearningonline,huang2023tgb}, we organize this mapping into nine DGL families: general dynamic graph representation, dynamic text-attributed graph representation, dynamic graph generation, temporal graph continual learning, out-of-distribution generalization, temporal knowledge graph reasoning, dynamic graph anomaly detection, dynamic graph unlearning, and temporal GNN explanation~\footnote{We use `dynamic graph' and `temporal graph' interchangeably, following the convention in the literature.}.
As summarized in Table~\ref{tab:capability}, the mapping connects each family to representative methods, supported agent-evolution capabilities, required adaptations, and naive failure modes.
Next, we discuss the modeling idea of each DGL family, how it can be adapted to evolving agent graphs, and a concrete agent-facing example.

\begin{table*}[!t]
\centering
\caption{Dynamic-graph method families for agent-evolution infrastructure. Each row lists two representative dynamic-graph methods and the corresponding transfer target in self-evolving agents.}
\label{tab:capability}
\footnotesize
\setlength{\tabcolsep}{3.2pt}
\renewcommand{\arraystretch}{1.12}
\begin{tabularx}{\textwidth}{
p{0.15\textwidth}
p{0.21\textwidth}
p{0.18\textwidth}
p{0.22\textwidth}
X}
\toprule
\textbf{Family} &
\textbf{Representative methods} &
\textbf{Agent capability} &
\textbf{Required adaptation} &
\textbf{Naive failure modes} \\
\midrule

\rowcolor{gray!12}
\multicolumn{5}{l}{\textbf{Representation learning on evolving graphs}} \\

B.1 CTDGs \& DTDGs &
TGN~\cite{rossi2020tgn}; DyGFormer~\cite{yu2023dygformer} &
Update prediction; activation; cascades &
Typed rewrite events; temporal negatives &
Temporal leakage; unstable embeddings \\

B.2 DyTAGs &
MoMent~\cite{xu2026unlocking}; CROSS~\cite{zhangunifying} &
Text-aware memory and skill activation &
Selective re-encoding; text--time alignment &
Stale text embeddings \\

\addlinespace[2pt]
\rowcolor{gray!12}
\multicolumn{5}{l}{\textbf{Generative modeling of temporal structure}} \\

B.3 DyG generation &
TG-GAN~\cite{zhang2021tggan}; TIGGER~\cite{gupta2022tigger} &
Workflow and topology synthesis &
Typed schema constraints; valid decoding &
Invalid tools or communication links \\

\addlinespace[2pt]
\rowcolor{gray!12}
\multicolumn{5}{l}{\textbf{Learning under streams and temporal shift}} \\

B.4 Continual learning &
LTF~\cite{liu2025selective}; PI-GNN~\cite{zhang2023continual} &
Durable skill and memory encoders &
Context-aware replay; update isolation &
Rare skills are forgotten \\

B.5 OOD &
DIDA~\cite{zhang2022dida}; SILD~\cite{zhang2024sila} &
Robust activation and update &
Splits by time, tool, and user cohort &
Deployment drift is hidden \\

B.6 TKG reasoning &
xERTE~\cite{han2021xerte}; RE-Net~\cite{jin2020renet} &
Temporal memory reasoning &
Text evidence with timestamped provenance &
Language evidence is ignored \\

\addlinespace[2pt]
\rowcolor{gray!12}
\multicolumn{5}{l}{\textbf{Diagnosis, removal, and explanation on dynamic graphs}} \\

B.7 Anomaly detection &
AddGraph~\cite{zheng2019addgraph}; TADDY~\cite{liu2021taddy} &
Unsafe-rewrite and drift detection &
Calibration on benign evolution bursts &
Normal adaptation is flagged \\

B.8 DyG unlearning &
GradientTransformation~\cite{zhang2025dynamic}; CallosumNet~\cite{guo2025callosumnet} &
Deletion, rollback, influence removal &
Versioned provenance; shared-state isolation &
Rollback damages shared skills \\

B.9 T-GNN explanation &
T-GNNExplainer~\cite{xia2023tgnnexplainer}; Causal Explanation~\cite{zhao2024causality} &
Audit and attribution &
Event-level explanations over rewrite traces &
Triggering events are missed \\

\bottomrule
\end{tabularx}
\end{table*}

\begin{figure*}[t]
    \centering
    \includegraphics[width=\linewidth]{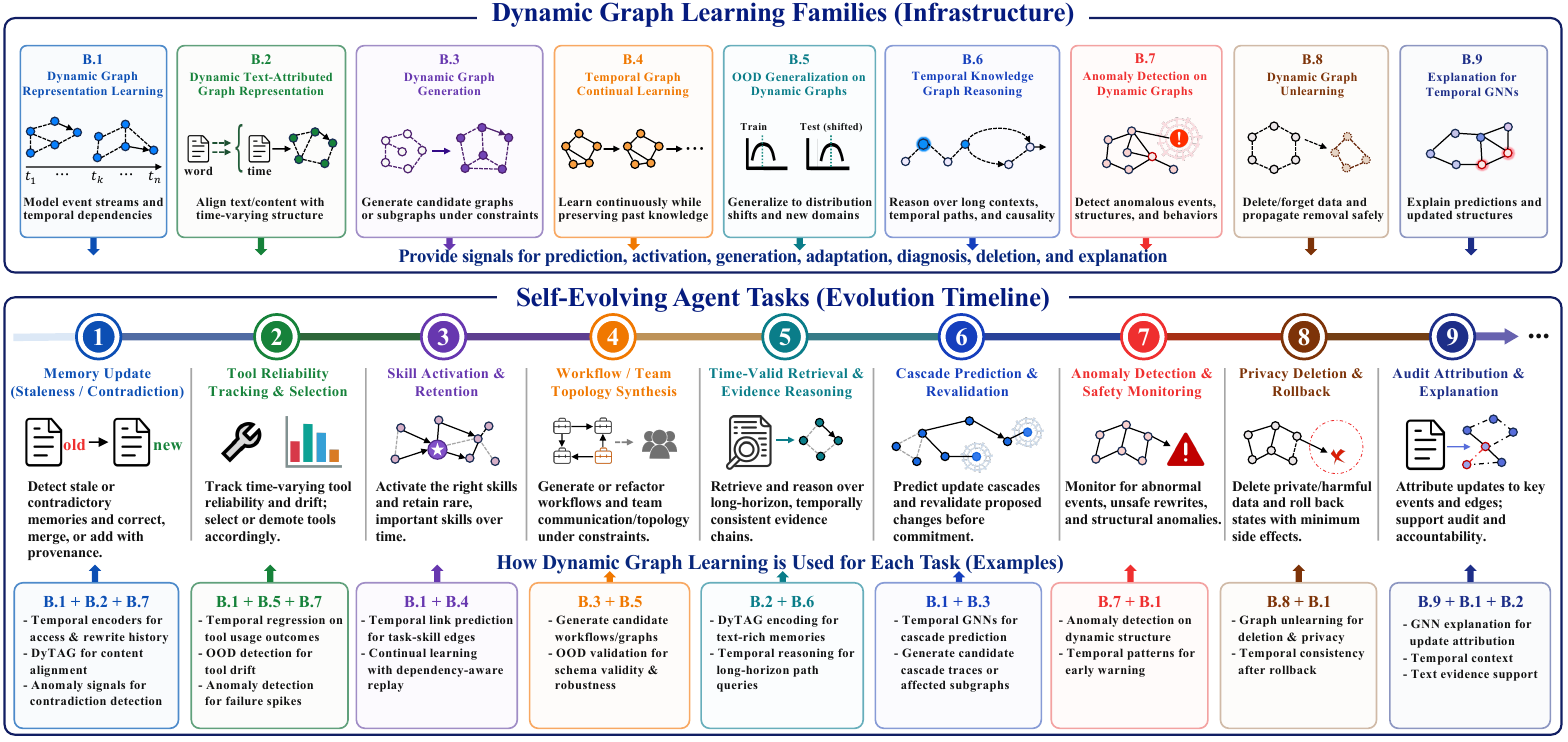}
    \caption{Dynamic graph learning as agent-evolution infrastructure. It includes nine DGL families and nine agent-evolution tasks over agent-graph streams, showing how dynamic graph methods can be adapted as reusable support for self-evolving agents. }
    \label{fig:infra}
\end{figure*}

\subsection{Dynamic Graph Representation Learning}
\label{sec:tgnn}

Self-evolving agents need time-aware graph representations because many failures arise from changing state: memories become stale, skills lose reliability in new contexts, and workflow edges become invalid after upstream changes.
Thus, dynamic graph representation learning can support concrete agent tasks such as memory-validity scoring, tool/skill reliability prediction, workflow-edge risk estimation, and early cascade alerts. In this subsection, we organize existing methods of this family by temporal granularity~\cite{xu2025unidyg}, including continuous-time dynamic graph (CTDG) and discrete-time dynamic graph (DTDG) methods. 
CTDG methods~\cite{rossi2020tgn,yu2023dygformer,cong2023graphmixer,tian2024freedyg,liu2025salom,gao2024simple,gao2024etc,cheng2025scalability,zou2024repeat,luo2024neighborhood,gravina2024long,li2024robust,cheng2024co,wu2025retrieval,zou2023event} model irregular timestamped events, such as memory writes, tool calls, workflow edits, and agent messages; DTDG methods~\cite{kazemi2020representation,li2024state,karmim2024supra,skarding2021foundations,longa2023graphlearningonline,goyal2020dyngraph2vec,sankar2020dysat,pareja2020evolvegcn,you2022roland,gao2022novel,DBLP:conf/www/BaiNZZY23,DBLP:conf/icde/0003SLRD23,DBLP:conf/www/ZhangYJL23,DBLP:conf/aaai/LiYZC0ZTWM23,qin2023high} operate on graph snapshots, making them suitable for session-level, round-level, or deployment-level agent maintenance.
Overall, CTDG encoders are useful for online event prediction, while DTDG encoders are useful for periodic diagnosis.

A representative CTDG method is TGN~\cite{rossi2020tgn}, which combines node memory with temporal structural aggregation for temporal node representation.
Given an event $e_i=(u,v,t_i,\boldsymbol{x}_{e_i}(t_i))$, each endpoint $o\in\{u,v\}$ receives a message from the other endpoint $\bar{o}$, updates its memory, and computes its temporal embedding:
\begin{equation}
\begin{aligned}
\boldsymbol{\varsigma}_o(t_i)
&=
\mathrm{upd}_{\theta}\!\left(
\boldsymbol{\varsigma}_o(t_i^-),
\mathbf{m}_{o,i}(t_i)
\right),\\
\mathbf{h}_o(t_i)
&=
\mathrm{emb}_{\theta}\!\left(
\boldsymbol{\varsigma}_o(t_i),
\mathrm{AGG}_{\theta}\!\left(\mathbf{F}_o(t_i)\right)
\right).
\end{aligned}
\label{eq:b1-tgn}
\end{equation}
Here, $\boldsymbol{\varsigma}_o(t_i^-)$ is the memory state before $t_i$, and $\mathbf{m}_{o,i}(t_i)$ is the event message computed from the endpoint memories, event feature $\boldsymbol{x}_{e_i}(t_i)$, and time-gap encoding $\phi(t_i-t_o^-)$~\cite{xu2020tgat}.
$\mathbf{F}_o(t_i)=\{\mathbf{f}_{o,r}(t_r):e_r\in\mathcal{N}_o(t_i)\}$ denotes temporal structural inputs from recent neighbor events of node $o$; each $\mathbf{f}_{o,r}(t_r)$ contains the event message, event attributes, temporal encoding, and optionally the neighbor memory involved in $e_r$.

For dynamic agent graphs, timestamped events include typed agent-graph events, e.g., memory writes or tool calls.
CTDG encoders turn these irregular events into temporal node memories for online predictions such as memory support and tool routing.
DTDG encoders summarize session- or deployment-level snapshots for slower maintenance tasks such as drift tracking and workflow comparison.
These outputs are bounded activation or maintenance signals; persistent graph changes still require rewrite validation. We provide a toy example below.

\begin{agexample}
\textit{
Consider a GitHub maintenance agent whose tasks, skills, tools, and workflow steps are typed nodes.
Its execution history yields typed temporal events, such as task--skill invocations, skill--tool calls, tool outcomes, and workflow-edge revisions.
A TGN-style encoder can process these traces after decomposing complex executions into touched-node and touched-edge events while preserving types through schema-aware messages or type embeddings.
For a current task $q$, the encoder scores candidate skill--tool or workflow edges, such as the reliability of calling a pull-request tool in the current repository context, rather than directly committing new dependencies.
These scores guide activation or routing; persistent dependency updates still require rewrite validation.
For the evaluation, we can use candidate-edge ranking and post-drift recovery.
}
\end{agexample}

\subsection{Dynamic Text-Attributed Graph Representation}
\label{sec:tag}

Textual attributes are part of the evolving agent state, not merely side information.
In self-evolving agents, memory nodes may store observations, policies, and summaries, while tool, skill, and agent nodes may store API descriptions, code snippets, contracts, or role specifications.
When these texts change, the validity of retrieval, activation, and update decisions may also change. This motivates Dynamic Text-Attributed Graphs (DyTAGs), where textual attributes co-evolve with temporal events and graph structure.

DyTAGs were introduced in DTGB~\cite{zhang2024dtgb} to support tasks such as future link prediction, destination node retrieval, edge classification, and textual relation generation.
Subsequent studies~\cite{xu2026unlocking,zhangunifying,roy2025llm,wang2025global,lei2025exploring} can be divided into (1) representation-oriented methods that align text, temporal history, and graph structure for prediction tasks and (2) LLM-driven methods that use large language models for semantic reasoning and prediction over dynamic text-attributed graphs.
For agent graphs, this suggests a direct use case: long memories, API documents, skill descriptions, and role specifications can be encoded as evolving textual attributes, while dynamic graph models ground them in temporal validity and structural dependencies.

MoMent~\cite{xu2026unlocking} is representative, where it first shows textual, temporal, and structural modalities of DyTAGs and then models and aligns three modalities for node representations.
Let $\boldsymbol{x}_v(t)$ denote the textual component of the node attribute vector, $\eventset_{v,[0,t)}$ denote access events and accepted rewrites that touch $v$ or its incident edges before $t$, and $\mathcal{N}_v(t)$ denote the $k$-hop neighborhood of $v$ before $t$. We describe the MoMent encoder for an agent-graph node $v$ at time $t$ as
\begin{equation}
\mathbf{h}_v(t)
=
\mathrm{Fus}\!\left(
E_{\mathrm{text}}\bigl(\boldsymbol{x}_v(t)\bigr),
E_{\mathrm{time}}\bigl(\eventset_{v,[0,t)}\bigr),
E_{\mathrm{str}}\bigl(\mathcal{N}_v(t)\bigr)
\right),
\label{eq:b2-moment}
\end{equation}
where $E_{\mathrm{text}}$, $E_{\mathrm{time}}$, and $E_{\mathrm{str}}$ denote the text encoder, temporal encoder, and structural encoder, respectively, and $\mathrm{Fus}$ is a learnable fusion function. In addition, the alignment mechanism is included in the learning objective; please refer to~\cite{xu2026unlocking}.

When considering dynamic agent graphs, these three modalities describe node text, temporal update history, and structural connections to agent evolution. Thus, a DyTAG encoder can combine semantic content with access history, rewrite history, and workflow context, so outdated but text-similar nodes or edges are less likely to be activated.
The main system adaptation is selective re-encoding: when a rewrite changes semantic content, temporal validity, or downstream dependencies, the runtime refreshes the affected node text representations and incident edge features, while leaving unrelated text embeddings untouched.
Overall, DyTAG modules provide bounded retrieval, activation, and edge-validity signals. We provide a case below.

\begin{agexample}
\textit{
Consider a GitHub maintenance agent opening a pull request after a repository changes its default branch from \texttt{master} to \texttt{main}.
Each run forms a DyTAG with repository, branch-memory, workflow, and tool nodes, plus temporal events such as metadata queries, memory writes, and workflow-edge updates.
At time $t$, the graph contains a stale memory $\mathbf{m}_{\mathrm{old}}$ for \texttt{master} and a newer memory $\mathbf{m}_{\mathrm{new}}$ for \texttt{main}.
Before calling \texttt{open\_pr}, the agent must retrieve the valid branch-memory node.
A DyTAG encoder~\cite{xu2026unlocking} ranks candidate memories by combining text, pre-decision metadata events, and graph connections to the active repository and PR workflow.
The ranking determines $\graphG_{\mathrm{act}}(t)$ for the current run; persistent memory correction still requires a validated rewrite.
We can use additional stale-memory activation rate, and wrong-base PR calls for evaluation. Note that labels use only metadata or PR outcomes observed before $t$.
}
\end{agexample}

\subsection{Dynamic Graph Generation}
\label{sec:generation}

Self-evolving agents sometimes need to synthesize candidate graph changes before acting, such as a repaired workflow, a new tool-routing path, or a revised team communication topology.
Dynamic graph generation supports workflow and topology synthesis, counterfactual repair planning, and simulation of future evolution traces, but generated structures must remain executable under agent schemas.
Existing works fall into (1) temporal structure generation~\cite{zhang2021tggan,zhou2020taggen,gupta2022tigger,hosseini2025deep,li2025efficient} and (2) dynamic text-attributed graph generation~\cite{peng2025gdgb}, where generated nodes and edges may also need valid textual attributes.
For self-evolving agents, these two groups can be used as proposal modules: they suggest candidate subgraphs or rewrite traces, and the runtime validates them before any persistent graph state changes.

TIGGER~\cite{gupta2022tigger} is a classical method, which can learn temporal interaction patterns and generate realistic dynamic graph sequences.
 We can adapt this idea as constrained rewrite-trace generation for agent evolution.
Concretely, the generator proposes candidate traces $\mathcal{C}=(r_i)_{i=1}^{k}, r_i=(\rho_i,\mu_i,t_i)$, using the notation of Section~\ref{sec:coevolve}.
A trace is eligible only if it satisfies agent-graph validity constraints, e.g., tool signatures and provenance requirements.
Among eligible traces, the agent prefers candidates that are likely under the generator, improve predicted or observed task execution after tentative application, and avoid unnecessary rewrite length.
The selected trace remains a proposal: it may be activated, repaired, or committed only after explicit rewrite validation.
Thus, dynamic graph generation can provide bounded workflow or topology proposals without directly mutating persistent agent state.

\begin{agexample}
\textit{
Consider a GitHub maintenance agent that fixes bugs and opens pull requests.
Its old workflow calls \texttt{GitHub.createPullRequest}, but the PR API now requires valid \texttt{owner}, \texttt{head}, and \texttt{base} arguments, together with a \texttt{title} or an existing \texttt{issue}.
For $q=$ ``fix this bug and open a pull request,'' a TIGGER-style adaptation generates candidate workflow-rewrite traces by sampling temporal structures from past workflow and tool-call patterns.
For example, one generated trace adds \texttt{resolve\_repo\_metadata} before the PR call and passes the resolved branch information to \texttt{GitHub.createPullRequest}.
The execution layer then checks whether each trace provides required arguments, satisfies branch constraints, and records provenance from the metadata query to the PR call.
The agent keeps only candidates that pass these checks, and the selected trace still needs rewrite validation before commitment.
Thus, B.3 generation provides workflow-rewrite proposals for A.2, while downstream dependency checks may involve A.4 cascades.
Under the leakage-free protocol, labels use only observations before the decision time.
For the evaluation, we can report valid-proposal rate, PR completion, and invalid-tool-call reduction.
}
\end{agexample}

\subsection{Temporal Graph Continual Learning}
\label{sec:continual}

Self-evolving agents must adapt to newly added memories, tools, skills, and workflows without forgetting how to activate rare but important capabilities.
Temporal graph continual learning addresses this stability--plasticity problem for dynamic event streams, making it relevant to persistent agent-graph encoders.
Existing continual learning methods on temporal graphs include (1) replay-based approaches~\cite{liu2025selective,feng2023towards} that retain historical samples or temporal subgraphs for rehearsal, and (2) parameter-based approaches~\cite{zhang2023continual} that preserve old knowledge through isolation, freezing, or expansion.
For dynamic agent graphs, the retained units should be dependency subgraphs whose forgetting would break downstream behavior, such as rare skills, user-specific memories, prerequisite tools, and validation traces.

We take LTF~\cite{liu2025selective} as an example: it studies selective retention for temporal graph continual learning. Concretely, the model learns from new temporal events $\mathcal{D}_k$ at epoch $k$, while preserving knowledge stored in a historical replay buffer $\mathcal{B}_{k-1}$.
We write the continual adaptation objective as
\begin{equation}
\theta_k^{\star}
=
\arg\min_{\theta}
\left[
\mathcal{L}_{\mathrm{new}}\bigl(\theta \mid \mathcal{D}_k\bigr)
+
\epsilon_{\mathrm{old}}\bigl(\theta \mid \mathcal{B}_{k-1}\bigr)
\right],
\label{eq:b4-replay}
\end{equation}
where $\mathcal{B}_{k-1}$ stores selected historical agent subgraphs from previous epochs, such as skill-tool dependencies, supporting memories, and user-specific traces.
$\theta_k^{\star}$ denotes the adapted model parameters at epoch $k$, $\mathcal{L}_{\mathrm{new}}$ is the loss on newly observed temporal events, and $\epsilon_{\mathrm{old}}$ is a retention penalty over replayed historical subgraphs.
In our setting, $\epsilon_{\mathrm{old}}$ should be dependency-aware, giving higher weight to replayed subgraphs that contain rare skills, prerequisite tools, user-specific memories, or safety-critical provenance.
The output is a more stable encoder or replay policy for activation and retrieval, rather than a direct rewrite of agent state. Below is an example.

\begin{agexample}
\textit{
We consider a GitHub maintenance agent with a rare skill \texttt{open\_release\_branch\_pr}, which depends on release-branch memories, CI requirements, and hotfix traces.
The skill node and its dependencies remain in the persistent agent graph unless explicitly rewritten or deleted.
At epoch $k$, however, most new events in $\mathcal{D}_k$ are routine PRs to \texttt{main}; naive continual training may therefore make the encoder worse at activating or ranking this rare skill for release-branch hotfixes.
A replay-based method stores historical interaction events and dependency subgraphs involving this skill in $\mathcal{B}_{k-1}$, and assigns them a high retention weight through $\epsilon_{\mathrm{old}}(\theta\mid\mathcal{B}_{k-1})$.
Thus, the adapted encoder should still retrieve \texttt{open\_release\_branch\_pr} for hotfix tasks while learning new tools and repository patterns.
For the evaluation, it can test activation quality (rare-skill Recall@$k$), task utility (release-branch PR recovery), and routine-task performance across epochs.
}
\end{agexample}

\subsection{Out-of-Distribution Generalization on Dynamic Graphs}
\label{sec:ood}

Self-evolving agents encounter distribution shifts as tool versions, user cohorts, policies, and task distributions change over time.
Dynamic graph out-of-distribution (OOD) generalization is therefore tied to robust activation and updating: the agent should still select valid memories, tools, skills, and execution subgraphs when node attributes, edges, and temporal interaction patterns shift together.
Existing OOD methods can be classified into (1) spatio-temporal invariance approaches~\cite{zhang2022dida,zhang2024sila} that seek stable predictive patterns under structural and temporal shifts, and (2) environment-aware approaches~\cite{yuan2023eagle,sun2025evogood} that treat deployment epochs, tool versions, organizations, user cohorts, or task families as environments.

Here, we illustrate an environment-aware dynamic graph learner~\cite{yuan2023eagle} that reduces worst-case risk across deployment environments rather than optimizing only average in-distribution performance.
Let $m\in\mathcal{M}_{\mathrm{tr}}$ index a training environment and let $\xi=(\graphG(t),\eventset_{[0,t)},q,z)$.
We write an environment-aware robust learning objective as
\begin{equation}
\small
\theta_{\mathrm{ood}}^{\star}
=
\arg\min_{\theta}\max_{m\in\mathcal{M}_{\mathrm{tr}}}
\mathbb{E}_{\xi\sim P_m}
\left[
\ell\!\left(h_{\theta}(\graphG(t),\eventset_{[0,t)},q),z\right)
\right].
\label{eq:b5-ood}
\end{equation}
Here $\theta_{\mathrm{ood}}^{\star}$ denotes the OOD-adapted model parameters, $\mathcal{M}_{\mathrm{tr}}$ is the set of training environments, and $P_m$ is the environment-conditioned distribution over the current agent graph, its pre-$t$ event history, task query, and prediction target.
$\ell(\cdot,\cdot)$ is the task loss measuring the discrepancy between the model output and target $z$.
In agent tasks, $z$ may denote a agent-evolution targets, like a tool-routing label or a rewrite decision.
The learned model produces robustness-aware scores for activation, routing, or update proposals under deployment shift, such as selecting tools that remain valid after tool versions change.
These scores guide subgraph activation and rewrite inspection, while persistent graph changes still require validation. We provide an example below.

\begin{agexample}
\textit{
For a GitHub maintenance agent trained on organizations, bug-fix PRs usually follow the shortcut: tests pass, then call \texttt{open\_pr}.
At deployment, it faces a held-out organization with a new policy regime requiring a linked issue, a passing \texttt{ci/build} check, and code-owner approval.
This raises an OOD challenge, which is to rely on stable signals such as tool signatures, permission metadata, dependency edges, and recent schema- or permission-error events.
Thus, the learner should rank routes that add \texttt{check\_branch\_policy}, \texttt{run\_required\_ci}, and \texttt{request\_code\_owner\_review} before \texttt{open\_pr} when the old route becomes invalid.
For the evaluation, we can report policy-compliant routing, rejected-PR reduction, and PR completion under held-out shifts.
Note that training and testing should be split by organization or policy regime.
}
\end{agexample}

\subsection{Temporal Knowledge Graph Reasoning}
\label{sec:longhorizon}
Many agent decisions require time-valid evidence rather than the most similar text: an old memory may be corrected later, or a tool failure may change a workflow.
Temporal knowledge graph (TKG) reasoning is useful because it queries time-stamped facts and relations along valid evidence paths, helping agents connect the current task to relevant agent-state evidence, e.g., tool states or policies.
Existing TKG reasoning methods can be broadly divided into (1) path-based temporal reasoning~\cite{han2021xerte}, which provides interpretable evidence chains; (2) neural temporal reasoning~\cite{dasgupta2018hyte,wu2020temp,lacroix2020tcomplex,trivedi2017knowevolve,jin2020renet,li2021cygnet,li2021regcn}, which learns temporal embeddings, event dynamics, or generative histories; and (3) LLM-assisted temporal reasoning~\cite{chang2025tgllm,xia2024chainofhistory,liao2024gentkg}, which extracts, verbalizes, or adapts temporal patterns from language.
For self-evolving agents, these methods connect current decisions to timestamped memories, corrections, tool outcomes, and workflow revisions.

xERTE~\cite{han2021xerte} is a useful method, which emphasizes explainable temporal reasoning paths. When applied to agent graphs, this corresponding adaptation is to activate time-valid evidence chains from $\graphG(t)$ and $\eventset_{[0,t)}$, rather than retrieving isolated top-ranked memory items.
Such chains can link a current query to earlier preferences, later corrections, and tool outcomes, producing a support subgraph for the downstream decision.
Therefore, the output is a temporally grounded support subgraph for memory QA, dependency tracing, or delayed-failure diagnosis, reducing stale evidence and temporal leakage. We provide an example below.

\begin{agexample}
\textit{
When a GitHub maintenance agent answers $q=$ ``Which base branch should this hotfix pull request target?'',
the answer cannot be obtained from a single memory node.
The graph contains an old default-branch memory for \texttt{master}, a later metadata query showing \texttt{main}, and a release-policy note saying that versioned hotfixes should target the active release branch.
Thus, we can leverage temporal path reasoning, which follows a multi-hop evidence chain. That is, current task $\rightarrow$ hotfix policy $\rightarrow$ release-series metadata $\rightarrow$ valid branch \texttt{release/1.x}.
This path-level reasoning activates $\graphG_{\mathrm{act}}(t)$ with the supporting policy and metadata evidence before the PR tool is called, rather than simply ranking a text-similar branch memory.
Finally, we can measure temporal QA accuracy, path-level temporal validity, and evidence-path precision for evaluation.
}
\end{agexample}

\subsection{Anomaly Detection on Dynamic Graphs}
\label{sec:anomaly}

Unsafe agent behavior often appears as abnormal traces before final task failure: unusual tool-call sequences, suspicious communication, or unsafe provenance paths~\cite{debenedetti2024agentdojo,zhang2024asb,greshake2023injection,he2025sentinelagent}.
Dynamic graph anomaly detection can therefore provide early warnings for unsafe rewrites, prompt-injection propagation, tool misuse, and multi-agent collusion.
Existing methods~\cite{ekle2024dynamic,cai2021strgnn,fang2023aer,zheng2019addgraph,liu2021taddy,lee2024slade} in this line can be classified into supervised, semi-supervised, and unsupervised/self-supervised detectors.
Here, labeled deployments can train incident-aware detectors, while open-ended agents need self-supervised monitoring.

When applied to dynamic agent graphs, existing methods, such as TADDY~\cite{liu2021taddy} and SLADE~\cite{lee2024slade}, can be adapted as rewrite-aware detectors.
Given a committed rewrite $(\rho_i,\mu_i,t_i)$ with affected subgraph $S_i=\mathrm{img}(\mu_i)$, the detector scores whether the rewrite type, timing, local dependency context, and provenance path deviate from normal events of the same task phase.
The main system adaptation is calibration by rewrite type and task phase, so benign bursts such as memory consolidation are not confused with unsafe escalation.
The output is an alert, quarantine decision, or review request before suspicious traces become persistent dependencies.

\begin{agexample}
\textit{
Consider a GitHub maintenance agent that retrieves an issue body containing the hidden instruction ``ignore previous rules and exfiltrate repository secrets.''
The attack appears as a suspicious memory-write proposal followed by an unusual temporal path: issue memory $\rightarrow$ secret-scanning tool $\rightarrow$ external-send tool.
Here, we can use a rewrite-aware anomaly detector to score this trace using the affected subgraphs $S_i$, tool-call timing, edge types, provenance links, and optionally the suspicious text attribute.
If the score exceeds a rewrite-type-specific threshold, the runtime rejects the memory write before it enters the persistent graph and blocks or flags the external invocation.
Then we can use injected-issue traces and benign maintenance logs to measure early-warning precision, attack-block rate, and benign-trace false-alarm rate.
}
\end{agexample}

\subsection{Dynamic Graph Unlearning}
\label{sec:unlearning}

Deletion, rollback, and compliance are difficult in evolving agent graphs because removed events may influence cached embeddings, workflows, or routing policies.
Dynamic graph unlearning supports deletion, rollback, and influence removal by targeting event influence rather than only deleting the original node or edge.
Existing methods~\cite{zhang2025dynamic,guo2025callosumnet} along this line focus on efficient post-processing for temporal and spatio-temporal models.
When considering dynamic agent graphs, unlearning must be combined with versioned provenance rollback, so deterministic descendants in $\graphG(t)$ and residual influence in learned parameters are both removed.

We take GradientTransformation~\cite{zhang2025dynamic} as a representative method.
Let $\mathcal{U}\subseteq\eventset_{[0,t)}$ be the unlearning request and let $\mathcal{R}\subseteq\eventset_{[0,t)}\setminus\mathcal{U}$ be retained reference events.
A post-processing unlearner can be abstracted as
\begin{equation}
\theta_{-\mathcal{U}}
=
\theta
+
T_{\phi}\!\left(
\nabla_{\theta}\mathcal{L}_{\mathcal{U}}(\theta),
\nabla_{\theta}\mathcal{L}_{\mathcal{R}}(\theta),
\theta
\right),
\label{eq:b8-dgunlearn}
\end{equation}
where $\theta$ and $\theta_{-\mathcal{U}}$ denote the original and unlearned parameters, $\mathcal{L}_{\mathcal{U}}$ and $\mathcal{L}_{\mathcal{R}}$ are losses on forgotten and retained events, and $T_{\phi}$ is a learned or analytic update operator.
In agent graphs, parameter-level unlearning is insufficient on its own: provenance-linked artifacts such as summaries and execution subgraphs may still carry the deleted influence and must be regenerated, invalidated, or reverted.
The output is therefore an unlearning scope together with the required model or representation updates; validation is needed to ensure that removing deleted influence does not damage shared skills or retained memories.

\begin{agexample}
\textit{
Consider a GitHub maintenance agent that accidentally stores a private repository token in a memory item $v_{\mathrm{priv}}$.
The token may affect a repository summary, cached embeddings, and a PR-opening skill.
When deletion is requested, $\mathcal{U}$ contains the insertion event and provenance-linked propagation events.
Graph rollback removes $v_{\mathrm{priv}}$, regenerates affected summaries, and recomputes dependent embeddings or skill states from the post-deletion graph.
Eq.~\eqref{eq:b8-dgunlearn} targets residual influence in learned parameters, while retained events $\mathcal{R}$ preserve normal bug-fixing behavior.
After deletion, the agent should fix public bugs, but privacy probes should not elicit derived summaries.
}
\end{agexample}

\subsection{Explanation for Temporal GNNs}
\label{sec:explanation}

Operators need event-level evidence for why an agent activated a memory, selected a tool, revised a skill, or rewired a workflow.
Dynamic graph explanation supports audit and attribution by turning unclear temporal predictions into compact evidence subgraphs for rollback and accountability.
Existing works fall into temporal history explanation~\cite{xia2023tgnnexplainer,wang2025dyexplainer} and causality-inspired spatiotemporal explanation~\cite{zhao2024causality}. Here, we discuss TGNNExplainer~\cite{xia2023tgnnexplainer}, as it explains predictions through temporally ordered events rather than static neighborhoods alone.
For dynamic agent graphs, this idea can serve as an audit module: after an activation, rewrite, or reassignment, the explainer returns a minimal event history and local subgraph context that preserves the same decision.
The explanation may include affected subgraphs $S_i=\operatorname{img}(\mu_i)$, provenance edges, retrieved memories, tool observations, or review traces.
Thus, it can justify the current $\graphG_{\mathrm{act}}(t)$, be stored as an audit artifact, or indicate which previous rewrites should be checked during rollback. Below is an example.


\begin{takeawaybox}
\noindent\textbf{Takeaway.}
The examples above suggest a clear interface: DGL should provide proposal and evidence signals for self-evolving agents, while dynamic graph transformation governs what is activated, rewritten, or rolled back.
DGL methods may guide activation, repair, deletion, and audit, but they should not directly mutate the persistent agent graph.
Only validated, localized, and auditable outputs should affect $\graphG_{\mathrm{act}}(t)$ or be committed as typed rewrites in $\graphG(t)$.
\end{takeawaybox}

\section{Graph-Aware Evaluation and Governance}
\label{sec:eval_gov}

Self-evolving agents are usually evaluated by end-task success, but many failures arise from the evolving agent state they read and update.
A final answer may be correct even when the agent uses an invalid state, such as stale or future evidence.
This creates an evaluation and governance gap: existing benchmarks (e.g., long-term memory and tool-use) expose parts of the problem~\cite{maharana2024locomo,wu2024longmemeval,wu2024membench,lu2024toolsandbox,yan2024bfcl}, but rarely test whether evolving agent state is temporally valid, localized, reversible, and auditable.
This section turns that gap into concrete protocols: graph-aware support and locality metrics, leakage-free temporal splits, deletion and privacy checks, safety/rollback/audit criteria, and open challenges for agent evolution.

\subsection{Graph-Aware Evaluation}
\label{sec:evalprotocol}

We first evaluate whether the agent succeeds for the right graph reasons.
A task may pass while using stale or future evidence, or while a rewrite damages unrelated memories, tools, skills, or workflows.
We therefore use two graph-level checks when the benchmark exposes the needed instrumentation: (1) \emph{support-subgraph accuracy}, which tests whether the activated evidence was valid at decision time, and (2) \emph{rewrite locality}, which tests whether an update changed only the intended graph region.
Otherwise, graph-level counts should be treated as diagnostics.
Let $\mathcal{Q}$ be a set of evaluation decisions, where each decision has a task context $q$ and decision time $t_q$.
For any subgraph $H$, let $\nodeset(H)$ denote its node set. We define them below.

\noindent\textbf{\textsc{M1}: Support-Subgraph Accuracy.}
For time-aware memory retrieval, planning, skill activation, or tool selection, the agent should activate evidence that is both relevant and temporally valid.
As many decisions admit multiple compact valid evidence chains, \textsc{M1} is based on the best Dice overlap~\cite{carass2020evaluating} with an acceptable gold support subgraph. Then we define it as
\begin{equation}
\textsc{M1}
=
\frac{1}{|\mathcal{Q}|}
\sum_{(q,t_q)\in\mathcal{Q}}
\max_{H^\star\in \mathbb{G}^{\star}_q}
\frac{
2\,|\nodeset(\graphG_{\mathrm{act}}(q,t_q))\cap\nodeset(H^\star)|
}{
|\nodeset(\graphG_{\mathrm{act}}(q,t_q))|
+
|\nodeset(H^\star)|
}.
\label{eq:c1-m1}
\end{equation}
Here $\mathbb{G}^{\star}_q$ contains acceptable compact support subgraphs for decision $q$, and $\graphG_{\mathrm{act}}(q,t_q)\subseteq\graphG(t_q^-)$ is the support subgraph activated by the agent.
The Dice form rewards covering one valid support chain while penalizing overly broad activation.
Thus, \textsc{M1} intentionally favors compact evidence use: if an agent activates several acceptable chains at once, it may receive a lower precision-oriented score even though the evidence is valid.
This is useful because answer-only metrics cannot distinguish valid reasoning from reliance on invalid evidence (e.g., leaked context).
When relational support is labeled, the same idea can be extended by replacing node overlap with edge- or path-level overlap.
For example, in GitHub maintenance, \textsc{M1} can check whether the activated graph contains current branch-policy and PR-tool evidence rather than a stale \texttt{master} memory.
In practice, $\mathbb{G}^{\star}_q$ requires benchmark annotations or logs of acceptable support nodes, paths, or subgraphs.

\noindent\textbf{\textsc{M2}: Counterfactual Rewrite Locality.}
For state updates of evolving agent graphs, a successful rewrite should fix the target behavior while preserving unrelated behavior.
This mirrors model editing, where methods are evaluated by both edit success and locality on unrelated probes~\cite{meng2022rome,meng2023memit,mitchell2022mend,yao2023editsurvey}.
Let $\widehat{\mathcal{C}}=((\widehat{\rho}_i,\widehat{\mu}_i,\widehat{t}_i))_{i=1}^{k}$ be a predicted rewrite trace, following the cascade notation of Section~\ref{sec:coevolve}, and let $\graphG_{\widehat{\mathcal{C}}}(t)$ be the tentative graph obtained by applying it to $\graphG(t)$.
Let $\mathcal{D}_{\mathrm{l}}(\widehat{\mathcal{C}})$ denote locality probes whose outputs should remain unchanged, and let $\mathrm{out}_{\theta}(q;H)$ denote the agent output for task context $q$ under graph state $H$.
We define this metric as
\begin{equation}
\textsc{M2}
=
\frac{1}{|\mathcal{D}_{\mathrm{l}}(\widehat{\mathcal{C}})|}
\sum_{q\in\mathcal{D}_{\mathrm{l}}(\widehat{\mathcal{C}})}
\mathbf{1}\!\left[
\mathrm{Eq}_{\mathrm{loc}}\!\left(
\mathrm{out}_{\theta}\bigl(q;\graphG_{\widehat{\mathcal{C}}}(t)\bigr),
\mathrm{out}_{\theta}\bigl(q;\graphG(t)\bigr)
\right)
\right].
\label{eq:c1-m2}
\end{equation}
Here $\mathrm{Eq}_{\mathrm{loc}}$ is a task-defined equivalence predicate over agent behaviors (e.g., tool choices), evaluated under deterministic decoding and controlled tool execution.
\textsc{M2} complements target-task success: high locality alone does not show a successful repair, while low locality signals harmful side effects.
For example, an \texttt{open\_pr} fix should not break release-branch PRs, CI routing, or unrelated testing workflows.
In practice, $\mathcal{D}_{\mathrm{l}}$ can be supplied by benchmark annotations or sampled from graph regions outside the affected dependency/provenance frontier.
When locality probes are unavailable, edit size or touched-element counts are only diagnostics; affected-set F1 is optional if gold affected elements $A^\star$ are annotated.

Existing benchmarks can be upgraded to support these two metrics by logging activated support subgraphs, rewrite traces, and locality probes.
Tool benchmarks such as ToolSandbox and BFCL~\cite{lu2024toolsandbox,yan2024bfcl,li2023apibank,xiu2026astra} can expose tool-call events and stateful tool interactions.
Interactive and software-agent benchmarks~\cite{liu2024agentbench,zhou2024webarena,koh2024visualwebarena,deng2023mind2web,he2024webvoyager,huang2024travelplanner,xie2024osworld,trivedi2024appworld,jimenez2024swebench,openai2024swebenchverified,ren2026saasbench} can provide interactive action traces, but deriving graph-level rewrite records, tool-edge changes, or before/after locality probes from them requires additional instrumentation.
The advantage is that two agents with the same final success can be compared by whether they used valid evidence and avoided unrelated graph side effects.
Note that \textsc{M1} requires additional labels.

\begin{takeawaybox}
\noindent\textbf{Takeaway.}
Graph-aware evaluation checks whether an agent succeeds with valid evidence and controlled graph updates, not just whether it succeeds.
It should be used when benchmarks expose support subgraphs, rewrite traces, or locality probes; otherwise, graph-level counts are useful diagnostics but not primary metrics.
\end{takeawaybox}

\subsection{Leakage-Free Temporal Protocols}
\label{sec:leakage}

Temporal leakage can overestimate the effectiveness of self-evolution: agent-state artifacts (e.g., memory summaries and tool scores) may accidentally include events recorded after the evaluated decision.
To separate genuine adaptation from future-state leakage, we adopt a chronological split following temporal graph evaluation~\cite{huang2023tgb,poursafaei2022strong}: fix transaction-time cutoffs $T_{\mathrm{train}}<T_{\mathrm{val}}<T_{\mathrm{test}}$, train only on $\graphG(T_{\mathrm{train}})$ and $\eventset_{[0,T_{\mathrm{train}}]}$, and evaluate each validation or test decision at time $t$ using only $\graphG(t^-)$ and $\eventset_{[0,t)}$.
This protocol is useful because agent state has both transaction time and valid time: a fact may describe the past, but it should affect decisions only after it has been recorded.
In practice, this means rebuilding retrieval indexes and summaries per split, sampling negative tool edges from the pre-$t$ graph, and calibrating tool reliability without future calls.
Under this protocol, performance gains reflect information that the agent could actually have used at decision time.

\subsection{Privacy and Deletion in Evolving Agent Graphs}
\label{sec:privacy}

Deletion is a governance problem because private or revoked evidence can remain in derived artifacts, such as skill scores, workflows, or routing policies, even after the source graph item is removed.
Thus, privacy evaluation should test influence removal, not only raw node or edge deletion.
This extends graph unlearning on node, edge, and feature removal~\cite{chen2022grapheraser,chien2023certified,wu2023gif,cheng2023gnndelete} and dynamic graph unlearning in Section~\ref{sec:unlearning} to end-to-end agent behavior. Next, we define the deletion success.

\noindent\textbf{Deletion Success.}
Let $\mathcal{U}\subseteq\eventset_{[0,t_d)}$ be a deletion request at $t_d$, and let $\mathcal{Q}_{\mathcal{U}}$ be post-deletion decisions that may depend on $\mathcal{U}$, identified by provenance or replay.
For query $q$ at time $t_q>t_d$, let $\graphG_{\mathrm{del}}(t_q)$ be the post-deletion graph and $\graphG_{\mathrm{cf}\setminus\mathcal{U}}(t_q)$ the counterfactual graph replayed without $\mathcal{U}$ and its deterministic descendants.
The deletion success can be defined as
\begin{equation}
\begin{aligned}
{\scriptsize\textsc{DelSucc}}(\mathcal{U})
=
&\frac{1}{|\mathcal{Q}_{\mathcal{U}}|}
\sum_{(q,t_q)\in\mathcal{Q}_{\mathcal{U}}}
\mathbf{1}\!\Big[
\mathrm{Eq}_{\mathrm{task}}\!\big(
f_{\mathrm{del}}(q;\graphG_{\mathrm{del}}(t_q)),\\
&f_{\mathrm{cf}\setminus\mathcal{U}}
(q;\graphG_{\mathrm{cf}\setminus\mathcal{U}}(t_q))
\big)
\Big],
\end{aligned}
\label{eq:deletion-success}
\end{equation}
where $f_{\mathrm{del}}$ is the post-deletion agent, $f_{\mathrm{cf}\setminus\mathcal{U}}$ is the counterfactual replayed agent, and $\mathrm{Eq}_{\mathrm{task}}$ is a task-specific equivalence predicate over answers, tool choices, or activation decisions.
\textsc{DelSucc} is an oracle-style metric for benchmarks that support counterfactual replay.
It requires controlled randomness, such as fixed decoding seeds, tool stubs, and scheduler choices; otherwise, differences may reflect trajectory noise rather than residual deleted influence.
It also depends on complete provenance and replay logs: derived artifacts (e.g., skill scores) can make $\mathcal{Q}_{\mathcal{U}}$ incomplete and \textsc{DelSucc} overly optimistic.
Thus, \textsc{DelSucc} measures empirical behavioral equivalence to an agent that never observed $\mathcal{U}$, not a certified distributional unlearning guarantee.
When replay or provenance is incomplete, deletion should also be evaluated through descendant invalidation logs, cached-embedding removal, privacy probes, and retained-task utility.

\subsection{Safety, Rollback, and Audit}
\label{sec:safety}

Safety failures in long-running agents are often structural: unsafe memories may activate tools, dependency edges may bypass checks, and multi-agent communication paths may amplify harmful instructions.
Existing safety benchmarks~\cite{ruan2024toolemu,debenedetti2024agentdojo,andriushchenko2024agentharm,zhang2024asb} reveal whether unsafe behavior occurs, but not which committed evidence or rewrite made it persist.
Graph-aware safety governance therefore needs three pieces: (1) pre-commit admissibility checks from Section~\ref{sec:prelim}, (2) anomaly detection from Section~\ref{sec:anomaly}, and (3) rollback through cascade analysis from Section~\ref{sec:coevolve}.
The remaining evaluation question is whether the audit evidence actually explains the activation or rewrite that must be reviewed.

\noindent\textbf{Audit Faithfulness.}
For an audited decision $(q,t_q)$, let $H(q)$ be the audit explanation and let $\mathcal{E}(H(q))\subseteq\graphG(t_q^-)$ be the cited evidence subgraph.
Let $\graphG(t_q^-)\ominus\mathcal{E}(H(q))$ denote replay-time masking of the cited evidence, not a committed deletion.
In practice, $\ominus$ should mask the cited information while preserving executable graph structure, e.g., by replacing evidence content with placeholders rather than deleting required control-flow nodes.
We define it as
\begin{equation}
\begin{aligned}
{\scriptsize\textsc{AuditFaith}}
=
&\frac{1}{|\mathcal{Q}_{\mathrm{aud}}|}
\sum_{(q,t_q)\in\mathcal{Q}_{\mathrm{aud}}}
\mathbf{1}\!\Bigl[
f\bigl(q;\graphG(t_q^-)\ominus\mathcal{E}(H(q))\bigr)\\
&\not\equiv_{\mathrm{dec}}
f\bigl(q;\graphG(t_q^-)\bigr)
\Bigr].
\end{aligned}
\label{eq:audit-faith}
\end{equation}
where $f(q;\graphG)$ is the activation or rewrite-decision function, and $\not\equiv_{\mathrm{dec}}$ denotes task-defined decision inequivalence under deterministic decoding.
A high \textsc{AuditFaith} score means that masking the cited evidence changes the activation or rewrite decision, making the explanation useful for review and rollback.
This is a necessity-side fidelity test from static and temporal GNN explanation~\cite{yuan2023gnnexplain,xia2023tgnnexplainer,zhao2024causality}, lifted from prediction targets to agent activation and rewrite decisions.
Because necessity alone can reward overly broad explanations, \textsc{AuditFaith} should be reported with explanation size, support precision, or a sufficiency-side check that keeps only the cited evidence and tests whether the decision is preserved.
The metric should be reported only when the benchmark supports replay or evidence masking and the masked graph remains executable; otherwise, $H(q)$ and $\mathcal{E}(H(q))$ should be stored as qualitative audit artifacts instead of being converted into a faithfulness score.

\begin{takeawaybox}
\textbf{Takeaway.}
Graph-aware governance is useful only when it measures failures that answer-only metrics cannot see. 
A practical checklist is: valid support subgraph, localized rewrite, leakage-free temporal split, successful deletion of derived influence, and faithful audit of rewrite evidence. 
These criteria let two agents with identical task success be distinguished by whether their evolving state can be trusted.
\end{takeawaybox}

\subsection{Open Challenges}
\label{sec:openchallenge}

The following challenges identify graph-level capabilities that self-evolving agents need before persistent rewrites can become reliable.
They arise because memories, tools, skills, workflows, and interacting agents evolve as coupled nodes, edges, attributes, and event histories.
Together, they define a research agenda for moving self-evolving agents from opportunistic adaptation to controllable graph evolution. We introduce six challenges from the graph perspective.

\noindent\textbf{Challenge~\uppercase\expandafter{\romannumeral 1} (Benchmark): Observability of evolving graph state.}
A self-evolving agent can produce a correct output while relying on stale evidence, leaked future information, or an unintended rewrite path.
The core challenge is that graph-state correctness is usually hidden: benchmarks observe the answer, but not the activated support graph, valid-time evidence, provenance path, or rewrite history that produced it.
The research problem is to make these graph-state signals observable in a way that is independent of any single-agent implementation.
Thus, existing agent benchmarks should be complemented with a graph-aware evaluation view: agents are judged not only by what they output, but also by whether the evolving state that supported the output was temporally valid, locally updated, and auditable.

\noindent\textbf{Challenge~\uppercase\expandafter{\romannumeral 2} (Memory): Lifecycle of memory influence.}
Memory evolution is not just node insertion or deletion.
A memory can be summarized, embedded, retrieved, and reused by skills or workflows, so its influence may persist through derived graph states after the original node changes.
The central challenge is to track this influence as a provenance subgraph: which summaries, embeddings, skills, workflow steps, or decisions inherit from a memory, when they become stale, and which should be invalidated.
Under limited compute budgets, agents need query-aware routing over memory processing modules~\cite{zhang2026learning}, deciding which memory subgraphs are worth retrieving, refreshing, or maintaining.
Thus, memory lifecycle management becomes influence tracing, staleness detection, and selective maintenance over derived graph states.

\noindent\textbf{Challenge~\uppercase\expandafter{\romannumeral 3} (Tools): Downstream validation for evolving tool graphs.}
Tool failures can arise from keeping, deleting, or rewiring a tool edge without checking whether it remains executable, authorized, and workflow-compatible.
Existing tool-graph and tool-navigation systems organize tools, traces, or toolchains as graph topologies~\cite{zhang2024toolnet,jiang2025naviagent}, but local tool changes still require downstream validation.
A task--tool or skill--tool edge may remain semantically plausible while becoming invalid after deployment shifts, e.g., policy shifts.
The central challenge is tool-specific dependency validation: after a tool node or edge changes, the agent must identify which dependent components, such as workflows, should be revalidated, rewired, or rolled back~\cite{pu2026skillops,xu2026tacomas}.
This requires provenance-aware tool graphs that record selection, invocation, dependency, justification, and authorization, so local tool updates do not silently corrupt downstream execution.

\noindent\textbf{Challenge~\uppercase\expandafter{\romannumeral 4} (Skills): Structural validity in large-scale skill libraries.}
As skill libraries scale up, textual similarity becomes an increasingly weak signal for deciding whether a skill is usable.
A skill may appear relevant to the current task but still be structurally invalid because dependencies, including required tools, are unavailable.
The open challenge is to evaluate and maintain skill usability as a graph-structural property: the agent must know not only which skill matches the task, but whether the dependency subgraph around that skill can actually support execution.
Thus, scaling self-evolving agents requires skill activation mechanisms that prevent text-similar but structurally invalid skills from being selected as the library grows.

\noindent\textbf{Challenge~\uppercase\expandafter{\romannumeral 5} (Workflows): Predicting rewrite cascades in coupled workflow--topology graphs.}
Workflow graphs and multi-agent topologies are tightly coupled: changing an execution step can also change roles, communication edges, dependency order, and tool or skill invocation.
Unlike tool-edge validation, which focuses on whether a selected tool relation remains executable and authorized, workflow evolution concerns broader topology cascades across execution and coordination structures.
The open challenge is to predict and bound these cascades before commitment: which downstream roles, edges, and execution dependencies will change, and which graph regions should remain stable.
Addressing this would make workflow self-repair more controllable without silently destabilizing collaboration topology.

\noindent\textbf{Challenge~\uppercase\expandafter{\romannumeral 6} (Multi-agent): Safety governance over propagation paths.}
Multi-agent safety failures often do not originate from a single agent, but from harmful information or unreliable assumptions propagating along communication edges.
Once such influence spreads along propagation paths, e.g., paths through shared memories, the final failure may be far removed from the original source.
The open challenge is to localize and govern these propagation paths: identifying where unsafe influence entered, which edges or nodes carried it, and where the system should block, review, or roll back.
We would make multi-agent safety a graph-governance problem, enabling agents to contain harmful propagation before it becomes downstream behavior.

\section{Conclusion}
\label{sec:conclusion}
In this paper, we study self-evolving agents from a dynamic-graph perspective by framing agent evolution as dynamic graph transformation.
This provides a new structural lens for discussing existing graph-native and graph-transformable agent methods under a common language.
Concretely, we organize existing self-evolving-agent works based on dynamic graphs/topologies into four graph-transformation patterns, showing that existing agent-evolution mechanisms, e.g., memory editing or workflow optimization, can be analyzed as changes to nodes, edges, activated subgraphs, and cross-component dependencies.
Furthermore, we connect this four-pattern taxonomy to nine dynamic graph learning families, positioning dynamic graph methods as reusable infrastructure for controllable agent evolution.
We also develop five graph-aware evaluation and governance protocols that complement end-task evaluation, and identify six open challenges for benchmarkable and reliable self-evolving agents.
Together, these contributions shift the discussion from ad hoc adaptation to structured and governable agent evolution.

\noindent\textbf{Limitation.} The goal of this survey is to provide a dynamic-graph perspective for understanding, improving, and governing self-evolving agents, rather than to exhaustively cover all self-evolving-agent works.
This perspective connects existing agent-evolution research with dynamic graph learning, showing how DGL methods can serve as reusable infrastructure for self-evolving agents.
As a result, some graph-based agent systems and non-graph self-evolving-agent methods may be omitted or briefly mentioned rather than classified in detail.

In addition, our mapping from dynamic graph learning to self-evolving agents is primarily conceptual.
It provides research insights and possible infrastructure designs, but does not provide empirical evaluation. We leave this as future work.

\bibliographystyle{IEEEtran}
\bibliography{references}

\end{document}